\documentclass[sigconf]{acmart}

\usepackage[utf8]{inputenc} 
\usepackage[T1]{fontenc}    
\usepackage{hyperref}       
\usepackage{url}            
\usepackage{booktabs}       
\usepackage{amsfonts}       
\usepackage{nicefrac}       
\usepackage{microtype}      
\usepackage{xcolor}         
\usepackage{comment}

\usepackage{microtype}
\usepackage{graphicx}
\usepackage{subfigure}
\usepackage{booktabs} 
\usepackage{amsfonts}
\usepackage{amsmath}
\usepackage{multicol}

\usepackage{pifont}
\usepackage{arydshln}

\usepackage{algorithm}
\usepackage{algorithmic}
\usepackage{hyperref}
\usepackage{cleveref}
\usepackage{xcolor}

\newcommand{\vc}[1]{\mathbf{#1}}
\newcommand{\mat}[1]{\mathbf{#1}}
\newcommand{\maskmatrix}{\mathcal{M}}
\newcommand{\maskmatrixat}[1]{\maskmatrix^{(#1)}}
\newcommand{\querymatrix}{\mat{X}}
\newcommand{\weightmatrix}{\mat{W}}
\newcommand{\weightmatrixat}[1]{\mat{W}^{(#1)}}
\newcommand{\chunk}{\mat{K}}
\newcommand{\chunkscalar}{K}
\newcommand{\weightmatrixchunk}[1]{\chunk^{(#1)}}
\newcommand{\predmatrix}{\mat{P}}
\newcommand{\predmatrixat}[1]{\mat{P}^{(#1)}}
\newcommand{\actmatrix}{\mat{A}}
\newcommand{\actmatrixat}[1]{\mat{A}^{(#1)}}
\newcommand{\featuredim}{d}
\newcommand{\querycount}{n}
\newcommand{\labelcount}{L}
\newcommand{\query}{\vc{x}}
\newcommand{\queryscalar}{x}
\newcommand{\weight}{\vc{w}}
\newcommand{\weightfull}[2]{\vc{w}^{(#1)}_{#2}}
\newcommand{\Rn}[1]{\mathbb{R}^{#1}}
\newcommand{\nzlist}[1]{\mathcal{S}(#1)}
\newcommand{\queryspace}{\mathcal{X}}
\newcommand{\labelspace}{\mathcal{Y}}
\newcommand{\clusterinst}{\tilde{\mathcal{Y}}}
\newcommand{\cluster}[2]{\mathcal{Y}^{(#1)}_{#2}}

\newcommand{\condpredmatrixat}[1]{\hat{\predmatrix}^{(#1)}}
\newcommand{\layer}{l}
\newcommand{\modelfunc}{f}
\newcommand{\labelinst}{\ell}
\newcommand{\beamedpredmatrix}{\tilde{\mat{P}}}
\newcommand{\beamedpredmatrixat}[1]{\beamedpredmatrix^{(#1)}}
\newcommand{\clusterind}{\mat{C}}
\newcommand{\clusterindat}[1]{\clusterind^{(#1)}}

\newcommand{\todo}[1]{{\color{red} TODO: #1}}

\definecolor{OliveGreen}{rgb}{0.1,0.5,0.1}

\AtBeginDocument{%
  \providecommand\BibTeX{{%
    \normalfont B\kern-0.5em{\scshape i\kern-0.25em b}\kern-0.8em\TeX}}}

\copyrightyear{2022} 
\acmYear{2022} 
\setcopyright{acmlicensed}\acmConference[WWW '22]{Proceedings of the ACM Web Conference 2022}{April 25--29, 2022}{Virtual Event, Lyon, France}
\acmBooktitle{Proceedings of the ACM Web Conference 2022 (WWW '22), April 25--29, 2022, Virtual Event, Lyon, France}
\acmPrice{15.00}
\acmDOI{10.1145/3485447.3511973}
\acmISBN{978-1-4503-9096-5/22/04}

\begin{document}

\title{Enterprise-Scale Search: Accelerating Inference for Sparse Extreme Multi-Label Ranking Trees}

\author{Philip A. Etter}
\email{paetter@stanford.edu}
\orcid{0002-6657-8230}
\affiliation{%
  \institution{Stanford University}
  \city{Stanford}
  \state{California}
  \country{USA}
}

\author{Kai Zhong}
\email{kaizhong@amazon.com}
\affiliation{%
  \institution{Amazon Search}
  \city{Berkeley}
  \state{California}
  \country{USA}
}

\author{Hsiang-Fu Yu}
\email{hsiangfu@amazon.com}
\affiliation{%
  \institution{Amazon Search}
  \city{Berkeley}
  \state{California}
  \country{USA}
}

\author{Lexing Ying}
\email{lexing@stanford.edu}
\affiliation{%
  \institution{Stanford University}
  \city{Stanford}
  \state{California}
  \country{USA}
}

\author{Inderjit Dhillon}
\email{isd@amazon.com}
\affiliation{%
  \institution{Amazon Search}
  \city{Berkeley}
  \state{California}
  \country{USA}
}

\renewcommand{\shortauthors}{Etter, Zhong, et al.}

\begin{abstract}
Tree-based models underpin many modern semantic search engines and recommender systems due to their sub-linear inference times. In industrial applications, these models operate at extreme scales, where every bit of performance is critical. Memory constraints at extreme scales also require that models be sparse, hence tree-based models are often back-ended by sparse matrix algebra routines. However, there are currently no sparse matrix techniques specifically designed for the sparsity structure one encounters in tree-based models for extreme multi-label ranking/classification (XMR/XMC) problems. To address this issue, we present the \emph{masked sparse chunk multiplication} (MSCM) technique, a sparse matrix technique specifically tailored to XMR trees. MSCM is easy to implement, embarrassingly parallelizable, and offers a significant performance boost to any existing tree inference pipeline at no cost. We perform a comprehensive study of MSCM applied to several different sparse inference schemes and benchmark our methods on a general purpose extreme multi-label ranking framework. We observe that MSCM gives consistently dramatic speedups across both the online and batch inference settings, single- and multi-threaded settings, and on many different tree models and datasets. To demonstrate its utility in industrial applications, we apply MSCM to an enterprise-scale semantic product search problem with 100 million products and achieve sub-millisecond latency of 0.88 ms per query on a single thread --- an 8x reduction in latency over vanilla inference techniques. The MSCM technique requires absolutely no sacrifices to model accuracy as it gives exactly the same results as standard sparse matrix techniques. Therefore, we believe that MSCM will enable users of XMR trees to save a substantial amount of compute resources in their inference pipelines at very little cost. Our code is publicly available at \url{https://github.com/amzn/pecos}, as well as our complete benchmarks and code for reproduction at \url{https://github.com/UniqueUpToPermutation/pecos/tree/benchmark}.
\end{abstract}




\maketitle

\section{Introduction}
\label{sec:intro}

Tree-based models are the workhorse of many modern search engines and recommender systems \cite{jasinska2020probabilistic, pecos2020,prabhu2018parabel,khandagale2020bonsai, wydmuch2018no, you2019attentionxml, pecos_semantic_kdd_2021}. Nonetheless, performing inference with these models on the scale demanded by modern applications quickly becomes intractable, requiring an unacceptable amount of compute time and memory to generate useful results. It is no surprise, then, that engineers devote a considerable amount of energy to optimizing their tree-based inference pipelines.

In the interest of pursuing tractable computation for practical applications, we dedicate this paper to examining how to make eXtreme Multi-label Ranking (XMR) and eXtreme Multi-label Classification (XMC)\footnotemark[1]\footnotetext[1]{We consider XMC as a subset of XMR. Everything in this paper regarding XMR applies equally well to XMC.} more time and memory efficient. In particular, we propose herein a set of optimizations for sparse XMR tree models that can drastically accelerate inference speed.

Most XMR tree models use a form of beam search to make inference tractable; the core idea of our optimizations is to take full advantage of the sparsity structure of this beam search to improve performance. This key idea manifests in our principal contribution, \emph{Masked Sparse Chunk Multiplication} (MSCM). MSCM is a new matrix data structure and accompanying multiplication algorithm that uses the aforementioned structured sparsity to drastically reduce unnecessary traversal and cache misses in the underlying memory. We observe that this technique can improve inference time usually anywhere from 2 to 20 times over vanilla inference.

The data structures and algorithms presented herein generalize to many different types of linear XMR tree models. In our performance benchmarks, we present an in-depth exploration of many variations of this technique implemented on top of our generic linear XMR tree implementation. 

We now quickly outline this paper: in \Cref{sec:rel_work}, we describe previous related work in the literature. In \Cref{sec:pecos}, we give an overview of a generic XMR tree model. \Cref{sec:ourmethod} includes all the details of our contribution to the literature. We present our performance benchmarks and evaluation in \Cref{sec:eval}. In our benchmarks, we compare the results of our methods to a baseline implementation without MSCM --- and demonstrate substantial improvements across a massive range of different settings, i.e., tree topologies, datasets, thread count, etc. We give advice on maximizing performance gain from MSCM in \Cref{sec:selectingmethod}. To demonstrate the usefulness of our method in an industrial setting, in \Cref{sec:enterprise} we apply our methods to an enterprise-scale semantic product search problem with 100 million products and use MSCM to achieve a sub-millisecond latency of 0.83 ms per query on a single thread --- an 8x improvement over the baseline. Finally, we conclude our study in \Cref{sec:conclusion}.

\section{Related Work}\label{sec:rel_work}

XMR tree models are widely used throughout the literature --- though a comprehensive terminology surrounding these related techniques has yet to fully emerge. In short, we define an XMR tree model as a tree where every node is associated with a ranker function, every label corresponds to a leaf of the tree, and the ranking for any given label is obtained by combining all ranker functions on the path from the label leaf to the root of the tree.

This definition captures several prior works --- for example, the probabilistic label tree (PLT) that originates from \cite{jasinska2016extreme}. This model in turn takes inspiration from previous work on label trees \cite{beygelzimer2009error, bengio2010label}, and there has since been a large body of follow-up work, spawning many state-of-the-art XMR models such as \textsc{Parabel} \cite{prabhu2018parabel}, \textsc{Bonsai} \cite{khandagale2020bonsai}, \textsc{ExtremeText} \cite{wydmuch2018no}, \textsc{AttentionXML} \cite{you2019attentionxml}, \textsc{NapkinXC} \cite{jasinska2020probabilistic}, and \textsc{PECOS} \cite{pecos2020}.  Note that \textsc{ExtremeText} and \textsc{AttentionXML} are designed only for dense features. The remaining models support sparse features and thereby require masked sparse matrix times sparse vector multiplication routines (see \cref{eq:maskedprediction}). We present MSCM as a way to accelerate these routines.

We turn now from the models themselves to existing optimizations for sparse inference. Unlike dense matrix multiplication, the random nature of memory access in sparse matrix algebra can be harder to optimize. Despite that, computational science researchers have devoted considerable energy to optimizing sparse matrix times vector products. Notable sparse matrix techniques relevant to this paper include \emph{cache blocking}, where the internal data of a matrix is reorganized into blocks to encourage memory locality. Many sparse kernel libraries make use of this idea to better tailor sparse matrix calculations to the underlying hardware. Notable examples include the SPARSITY framework \citep{im2004sparsity}, a framework that can automate the optimization of sparse kernels -- making use of both register and cache level optimizations. Other examples include work by \citep{williams2007optimization} specifically targeting multi-core platforms, and \citep{nishtala2007cache}, who develop an analytic model to better help predict optimal sparse kernel reorganization. Unfortunately, these techniques all target sparse matrix times dense vector calculations, as these are most common in computational science.

In comparison, sparse matrix times sparse vector (SpMSpV) multiplication is an under-served area. However, there are a number of emerging techniques for sparse matrix operations with underlying graph structure, including \cite{sundaram2015graphmat, azad2017work, yang2015fast}. Unfortunately, none of these methods are tailored to XMR tree inference, where beam search induces a very well-structured mask over matrix outputs. They are also significantly more heavy-weight than the method we present. The scale of XMR problems therefore necessitates the development of new masked sparse matrix methods specifically tailored to XMR trees.

\section{XMR Tree Models}\label{sec:pecos}

In this section, we present a generic formulation of the tree model for eXtreme Multi-label Ranking (XMR) problems. To provide the necessary context, we will give a brief overview of the inference algorithm. We omit training details because they are not directly relevant to the techniques in this paper once a model is trained; but we recommend that readers see any of the aforementioned XMR tree model papers e.g. \cite{prabhu2018parabel, khandagale2020bonsai, pecos2020} for an overview of training. 

\subsection{Overview} \label{sec:pecos_overview}

An XMR problem can be characterized as follows: given a query $\query$ from some embedding $\Rn{\featuredim}$ and a set of labels $\labelspace$, produce a model that gives an (implicit) ranking of the relevance of the labels in $\labelspace$ to query $\query$. In addition, for any query $\query \in \queryspace$, one must be able to efficiently retrieve the top $k$ most relevant labels in $\labelspace$ according to the model --- noting that $d$ is typically very large and $\query$ very sparse. 

\begin{figure}
\begin{center}
\includegraphics[scale=0.25]{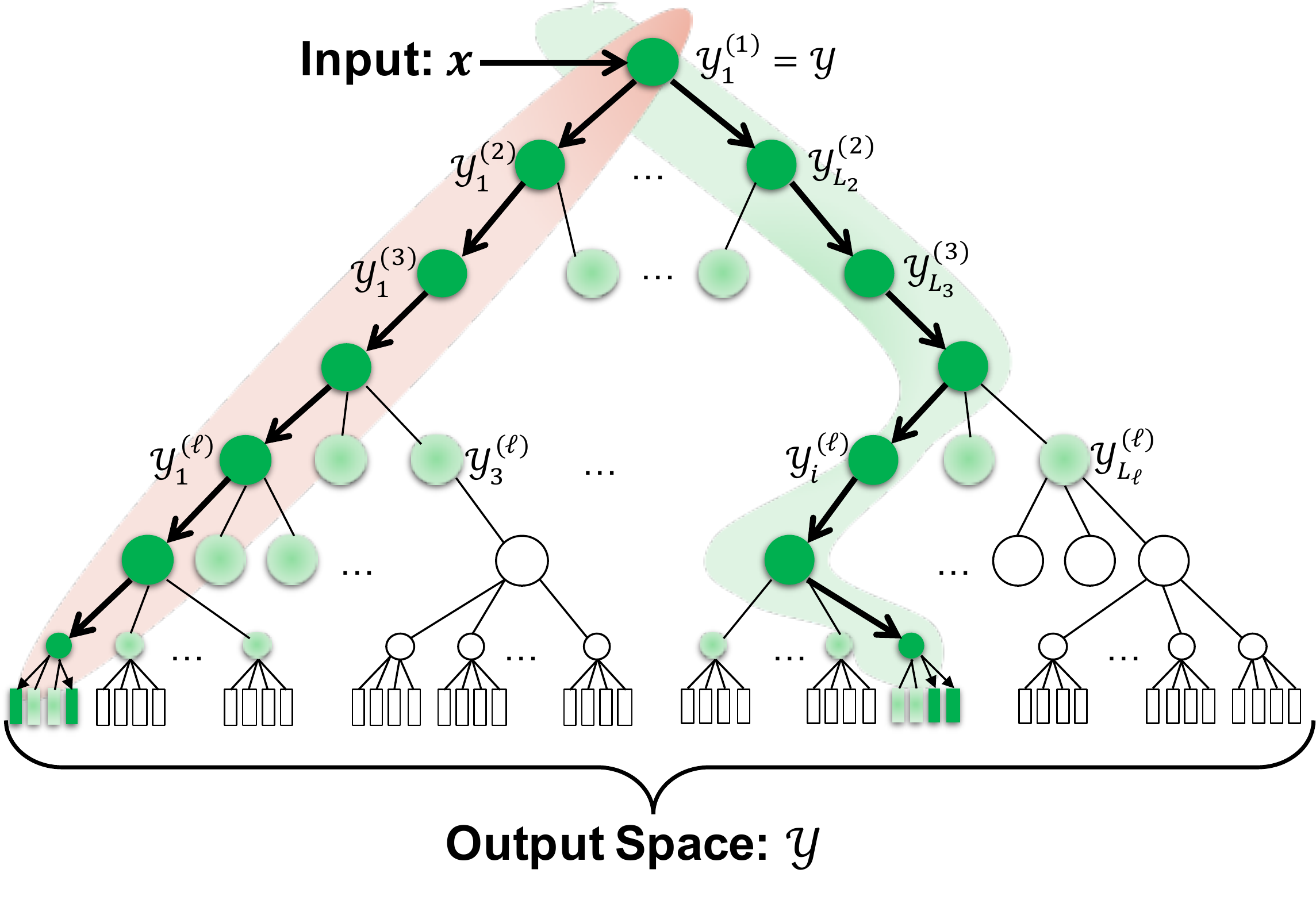}
\caption{A diagram of the structure of an XMR tree model. \cite{pecos2020}} \label{fig:pecosstruct}
\end{center}
\end{figure}

\begin{algorithm}
{ \scriptsize
   \caption{Linear XMR Tree Inference}
   \label{alg:batchinference}
\begin{algorithmic}[1]
   \STATE {\bfseries Input:} Query matrix $\querymatrix \in \Rn{\querycount \times \featuredim}$.
   \STATE {\bfseries Output:} Beamed predictions $\beamedpredmatrix \in \Rn{\querycount \times \labelcount}$.
   \STATE Initialize $1$st layer predictions to $1$: $\beamedpredmatrixat{1} = \mat{1} \in \Rn{\querycount \times 1}$.
  \FOR{$\layer \in \{2, 3, ..., \text{depth} \}$}
  \STATE Copy previous layer predictions:
  $\beamedpredmatrixat{\layer} \gets \beamedpredmatrixat{\layer - 1} {\clusterindat{\layer - 1}}^T\,.$
  \STATE Get computation mask: $\maskmatrixat{\layer} \gets \text{bool}(\beamedpredmatrixat{\layer} \neq 0)$.
    \STATE Conditional prediction step: $\condpredmatrixat{\layer} \gets \sigma(\maskmatrixat{\layer} \odot (\querymatrix \weightmatrixat{\layer}))$.
  \STATE Combine with prev. layers: $\beamedpredmatrixat{\layer} \gets \condpredmatrixat{\layer} \odot \beamedpredmatrixat{\layer - 1}$.
  \STATE Beam search by selecting top $b$ entries of each row:
  $\beamedpredmatrixat{\layer} \gets \text{SelectTop}_b(\beamedpredmatrixat{\layer}) \,.$
  \ENDFOR
   \STATE {\bfseries return} $\beamedpredmatrix = \beamedpredmatrixat{\text{depth}}$
\end{algorithmic}
}
\end{algorithm}

A linear XMR tree model is a hierarchical linear model that constructs a hierarchical clustering of the labels $\mathcal{Y}$, forming a tree structure. These clusters are denoted $\cluster{\layer}{i}$, where $\layer$ denotes the depth (i.e., layer) of $\cluster{\layer}{i}$ in the model tree and $i$ denotes the index of $\cluster{\layer}{i}$ in that layer, see \Cref{fig:pecosstruct}. The leaves of the tree are the individual labels of $\labelspace$.

Every layer of the model has a ranker model that scores the relevance of a cluster $\cluster{\layer}{i}$ to a query $\query \in \queryspace$. This ranker model may take on different forms, but for this paper we assume that the model is logistic-like. This means that, at the second layer, for example, the relevance of a cluster $\cluster{2}{i}$ is given by
\begin{equation}
\modelfunc(\query, \cluster{2}{i}) = \sigma\left({\weightfull{2}{i}} \cdot \query \right) \,,
\end{equation}
where $\sigma$ denotes an activation function (e.g., sigmoid) and $\weightfull{2}{i} \in \Rn{\featuredim}$ denotes a very sparse vector of weight parameters.

At subsequent layers, rankers are composed with those of previous layers, mimicking the notion of conditional probability; hence the score of a cluster $\clusterinst \subset \labelspace$ is defined by the model as
\begin{equation} \label{eq:defmodel}
f(\query, \clusterinst) = \prod_{\weight \in \mathcal{A}(\clusterinst)} \sigma \left( \weight \cdot \query \right) \,,
\end{equation}
where $\mathcal{A}(\clusterinst) = \{ \weightfull{\layer}{i} \mid \clusterinst \subset \cluster{\layer}{i}, \layer \neq 1 \}$ denotes all tree nodes on the path from $\clusterinst$ to the root $\labelspace$ (including $\clusterinst$ and excluding $\labelspace$). Naturally, this definition extends all the way to the individual labels $\labelinst \in \labelspace$ at the bottom of the tree. We assume here for simplicity that the leaves of the tree all occur on the same layer, but the techniques described in this paper can be extended to the general case.

As a practical aside, the column weight vectors $\weightfull{\layer}{i}$ for each layer $\layer$ are stored in a $\featuredim \times \labelcount_\layer$ weight matrix
\begin{equation}
\weightmatrixat{\layer} = \left[\begin{array}{cccc} \weightfull{\layer}{1} & \weightfull{\layer}{2} & ... & \weightfull{\layer}{\labelcount_\layer} \end{array}\right] \,,
\end{equation}
where $\labelcount_\layer$ denotes the number of clusters in layer $\layer$. The tree topology at layer $\layer$ is usually represented using a cluster indicator matrix $\clusterindat{\layer}$. $\clusterindat{\layer} \in \{0, 1\}^{\labelcount_{\layer + 1} \times \labelcount_{\layer}}$ is defined as
\begin{equation}
\clusterindat{\layer}_{ij} = \text{bool}(\cluster{\layer+1}{i} \subset \cluster{\layer}{j})\,,
\end{equation}
i.e., it is one when $\cluster{\layer+1}{i}$ is a child of $\cluster{\layer}{j}$ in the tree. Here, $\text{bool}(\cdot)$ is $1$ when the condition $\cdot$ is true and $0$ otherwise.

\subsection{Inference} \label{sec:batchinference}

Throughout this paper, we will assume two inference settings:

\begin{enumerate}
   \item \emph{Batch Inference}: inference is performed for a batch of $\querycount$ queries represented by a sparse matrix $\querymatrix \in \Rn{\querycount \times \featuredim}$ where every row of $\querymatrix$ is an individual query $\query_i$.
   \item \emph{Online Inference}: a subset of the batch setting where there is only one query, e.g., the matrix $\querymatrix$ has only one row.
\end{enumerate}

When performing inference, the XMR model $\modelfunc$ prescribes a score to all query-cluster pairs $(\query_i, \cluster{\layer}{j})$. Hence, in the batch setting, one can define the prediction matrices,
\begin{equation}
\predmatrixat{\layer}_{ij} = f(\query_i, \cluster{\layer}{j}) = \prod_{\weight \in \mathcal{A}(\cluster{\layer}{j})} \sigma \left( \weight \cdot \query_i \right)  \,.
\end{equation}
The act of batch inference entails collecting the top $k$ most relevant labels (leaves) for each query $\query_i$ and returning their respective prediction scores $\predmatrixat{\layer}_{ij}$.

However, the act of exact inference is typically intractable, as it requires searching the entire model tree. To sidestep this issue, models usually use a greedy beam search of the tree as an approximation. For a query $\query$, this approach discards any clusters on a given level that do not fall into the top $b \geq k$ most relevant clusters examined at that level. Hence, instead of $\predmatrixat{\layer}$, we compute {\em beamed} prediction matrices $\beamedpredmatrixat{\layer}$, where each row has only $b$ nonzero entries whose values are equal to their respective counterparts in $\predmatrixat{\layer}$. Pseudo-code for the inference algorithm is given for reference in \Cref{alg:batchinference}, where $\odot$ denotes entry-wise multiplication.

\section{Our Method} \label{sec:ourmethod}
Our contribution is a method of evaluating masked sparse matrix multiplication that leverages the unique sparsity structure of the beam search to reduce unnecessary traversal, optimize memory locality, and minimize cache misses. The core prediction step of linear XMR tree models is the evaluation of a masked matrix product, i.e.,
\begin{equation} \label{eq:maskedprediction}
\actmatrixat{\layer} = \maskmatrixat{\layer} \odot (\querymatrix \weightmatrixat{\layer}) \,,
\end{equation}

where $\actmatrixat{\layer} \in \Rn{\querycount \times \labelcount_\layer}$ denotes ranker activations at layer $\layer$, $\maskmatrixat{\layer} \in \{0, 1\}^{\querycount \times \labelcount_\layer}$ denotes a dynamic mask matrix determined by beam search, $\querymatrix \in \Rn{\querycount \times \featuredim}$ is a sparse matrix whose rows correspond to queries in the embedding space, $\weightmatrixat{\layer} \in \Rn{\featuredim \times \labelcount_\layer}$ is the sparse weight matrix of our tree model at layer $\layer$, and $\odot$ denotes entry-wise multiplication. Note the mask matrix $\maskmatrixat{\layer}$ is only known at the time of reaching the $l$-th layer. We leave out the application of $\sigma$ because it can be applied as a post processing step. We have observed that this masked matrix multiplication takes up the vast majority of inference time  on various data sets --- between $90\%$ and $98\%$ depending on model sparsity --- and hence is ripe for optimization.

\textbf{For readability, we will suppress the $(\layer)$ superscript for the rest of the section.} Recall that both $\weightmatrix$ and $\querymatrix$ are highly sparse. One typically implements the above computation by storing the weight matrix $\weightmatrix$ in compressed sparse column (CSC) format and the query matrix $\querymatrix$ in compressed sparse row (CSR) format\footnote{For details of the well-known CSR and CSC formats, see \cite{davis2006direct}.}, facilitating efficient access to columns of $\weightmatrix$ and rows of $\querymatrix$, respectively. Then, to perform the operation in \Cref{eq:maskedprediction}, one iterates through all the nonzero entries $\mathcal{M}_{ij} \neq 0$, extracts the $i$-th query $\query_i$ and the ranker weights $\weight_j$ for cluster $j$, and computes the inner product $\actmatrix_{ij} = \query_i \cdot \weight_j$ via a sparse vector dot product routine. One method is to use a progressive binary search in both $\query_i$ and $\weight_j$ to find all simultaneous nonzeros. We give pseudo-code for binary search in \Cref{alg:sparsevectorinnerprod}. However, there are other iteration methods, we will describe below -- such as marching pointers, hash-maps, and dense lookups. Any iteration method can be combined with MSCM to give a significant performance boost.

\begin{figure}
\centering
  \includegraphics[width=0.475\textwidth]{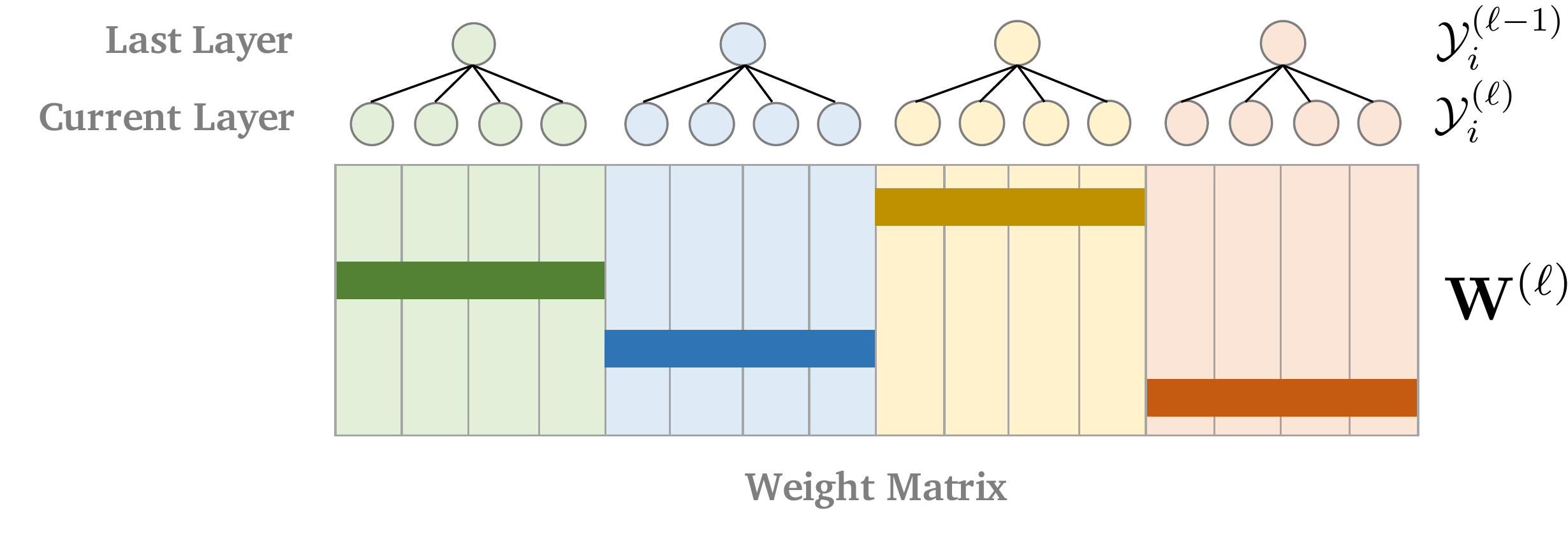}
  \hspace{0.1in}
  \includegraphics[width=0.475\textwidth]{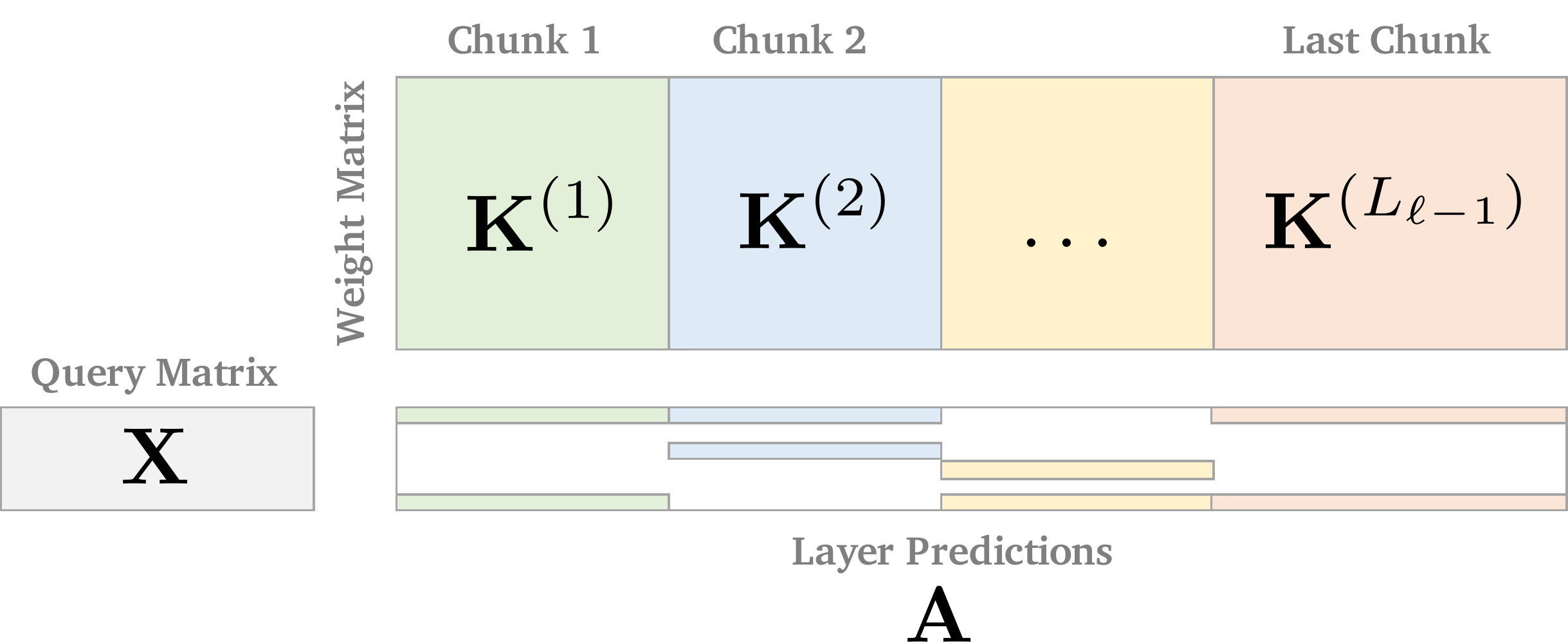}
  \caption{\textbf{Top}: A simplified pictorial representation of the property discussed in \Cref{it:similarsupport}. The darkened region indicate non-zeros and the light regions indicate zeros. \textbf{Bottom}: A visual representation of the chunked matrix multiplication routine. The colored blocks of $\actmatrix$ represent possible nonzeros. For each colored block in the output, MSCM evaluates the entire block simultaneously. Blocks are evaluated in order of their color.} \label{fig:explanation}
\end{figure}

While for general sparse matrix multiplication, this is a reasonable method of computation, there are a few special properties about the matrices $\maskmatrix$ and $\weightmatrix$ that this method ignores, namely:

\begin{enumerate}
\item The sparsity pattern of the mask matrix $\maskmatrix$ is determined by prolongating the search beam from layer $\layer - 1$ to layer $\layer$. That is, if cluster $\cluster{\layer}{j}$ in layer $l$ is a child of a cluster in the search beam of query $i$, then $\mathcal{M}_{ij} \neq 0$. This means that, if the nodes on a layer are grouped by their respective parents, the nonzero entries of $\mathcal{M}$ come in contiguous single row blocks corresponding to siblings in the model tree. See \Cref{fig:explanation}, bottom, for an illustration. \textbf{This suggests that one may improve performance by evaluating all siblings simultaneously.} \label{it:sparsityofmask}
\item Columns in $\weightmatrix$ corresponding to siblings in the model tree often tend to have \emph{similar sparsity patterns}. We give an illustration of this fact in \Cref{fig:explanation}, top. Together with the previous observation, \textbf{this suggests storing the matrix $\weightmatrix$ in such a way that entries of $\weightmatrix$ are contiguous in memory if they both lie on the same row and belong to siblings in the model tree.} \label{it:similarsupport}
\end{enumerate}

These considerations lead us naturally to the \textbf{column chunked matrix} data structure for the weight matrix $\weightmatrix$. In this data structure, we store the matrix $\weightmatrix \in \Rn{d \times L_l}$ as a horizontal array of \emph{matrix chunks} $\weightmatrixchunk{i}$,
\begin{equation} \label{eq:weightchunked}
    \weightmatrix^{(l)} = \left[\begin{array}{cccc} \weightmatrixchunk{1} & \weightmatrixchunk{2} & ... & \weightmatrixchunk{\labelcount_{\layer - 1}} \end{array} \right] \,,
\end{equation}
where each chunk $\weightmatrixchunk{i}\in \Rn{d\times B}$ ($B$ is the branching factor, i.e. number of children of the parent node) and is stored as a vertical sparse array of some sparse horizontal vectors $\vc{v}^{(j,i)}$,
\begin{equation}
\weightmatrixchunk{i} = \left[\begin{array}{ccccccc} \vc{0} & \hdots & (\vc{v}^{(r_1, i)})^T & \hdots & (\vc{v}^{(r_{s_i}, i)})^T & \hdots & \vc{0} \end{array}\right]^T \,.
\end{equation}
We identify each chunk $\weightmatrixchunk{i}$ with a parent node in layer $\layer- 1$ of the model tree, and the columns of the chunk $\weightmatrixchunk{i}$ with the set of siblings in layer $\layer$ of the model tree that share the aforementioned parent node in layer $\layer - 1$. As we will see, this data structure therefore makes use of both ideas \Cref{it:sparsityofmask} and \Cref{it:similarsupport} simultaneously.

To see why this data structure can accelerate the masked matrix multiplication, consider that one can think of the mask matrix $\maskmatrix$ as being composed of blocks,
\begin{equation}
\maskmatrix = \left[\begin{array}{cccc} \maskmatrix^{(1, 1)} & \maskmatrix^{(1, 2)} & ... & \maskmatrix^{(1, \labelcount_{\layer - 1})} \\ \vdots & \vdots & \ddots & \vdots \\ \maskmatrix^{(\labelcount_{\layer}, 1)} & \maskmatrix^{(\labelcount_{\layer}, 2)} & ... & \maskmatrix^{(\labelcount_{\layer}, \labelcount_{\layer - 1})} \end{array}\right] \,,
\end{equation}
where the block column partition is the same as that in \Cref{eq:weightchunked}, and every block has one row and corresponds to a single query. As per the observation in \Cref{it:sparsityofmask} above, every block $\maskmatrix^{(j, i)}$ must either be composed entirely of zeros or entirely of ones.

Therefore, since $\actmatrix$ and $\maskmatrix$ share the same sparsity pattern, the ranker activation matrix $\actmatrix$ is also composed of the same block partition as $\maskmatrix$,
\begin{equation}
\begin{split}
\actmatrix &= \left[\begin{array}{cccc} \actmatrix^{(1, 1)} & \actmatrix^{(1, 2)} & ... & \actmatrix^{(1, \labelcount_{\layer - 1})} \\ \vdots & \vdots & \ddots & \vdots \\ \actmatrix^{(\labelcount_{\layer}, 1)} & \actmatrix^{(\labelcount_{\layer}, 2)} & ... & \actmatrix^{(\labelcount_{\layer}, \labelcount_{\layer - 1})} \end{array}\right] \,, \\
\actmatrix^{(j, i)} &= \maskmatrix^{(j, i)} \odot (\query_j \weightmatrixchunk{i}) \,.
\end{split}
\end{equation}
Hence, for all mask blocks $\maskmatrix^{(j, i)}$ that are $1$, we have
\begin{equation} \label{eq:expforblock}
\actmatrix^{(j, i)} = \query_j \weightmatrixchunk{i} = \sum_{k \in S(\query_j) \cap S(\weightmatrixchunk{i})} \queryscalar_{jk} \vc{v}^{(k, i)} \,,
\end{equation}
where $S(\query_j)$ and $S(\weightmatrixchunk{i})$ denote the indices of the nonzero entries of $\query_j$ and the nonzero rows of $\weightmatrixchunk{i}$ respectively. The above equation says that for all entries of $\actmatrix$ in the same block, the intersection $k \in S(\query_j) \cap S(\weightmatrixchunk{i})$ only needs to be iterated through \emph{once per chunk}, as opposed to \emph{once per column} as is done in a vanilla implementation. Of course, it is theoretically possible that $S(\query_j) \cap S(\weightmatrixchunk{i})$ is substantially larger than the intersections $S(\query_j) \cap S(\weight_i)$ in our baseline, but this tends not to be the case in practice because of the observation in \Cref{it:similarsupport} that the columns of $\weightmatrixchunk{i}$ tend to have similar support. Moreover, the actual memory locations of the values actively participating in the product \Cref{eq:expforblock} are physically closer in memory than they are when $\weightmatrixchunk{i}$ is stored in CSC format. This helps contribute to better memory locality.

We take a moment to pause and remark that the only task remaining to fully specify our algorithm is to determine how to efficiently iterate over the nonzero entries $\queryscalar_{jk}$ and nonzero rows $\weightmatrixchunk{k, i}$ for $k \in \nzlist{\query_j} \cap \nzlist{\weightmatrixchunk{i}}$. This is essential for computing the vector-chunk product $\query_j \weightmatrixchunk{i}$ efficiently. There are number of ways to do this, each with potential benefits and drawbacks:

\begin{enumerate}
    \item \textbf{Marching Pointers}: The easiest method is to use a marching pointer scheme to iterate over $\queryscalar_{jk}$ and $\weightmatrixchunk{k,i}$ for $k \in \nzlist{\query_j} \cap \nzlist{\weightmatrixchunk{i}}$. In this scheme, we maintain both $\nzlist{\query_j}$ and $\nzlist{\weightmatrixchunk{i}}$ as sorted arrays. To iterate, we maintain an index $k_x$ in $\nzlist{\query_j}$ and an index $k_K$ in $\nzlist{\weightmatrixchunk{i}}$. At any given time, we either have $k_x = k_K$, in which case, we emit $\queryscalar_{jk}$ and $\weightmatrixchunk{k,i}$; if $k_x < k_K$, we increment $k_x$; and if $k_x > k_K$, we increment $k_x$. \label{enum:march}
    \item \textbf{Binary Search}: Since $\query$ can be highly sparse, the second possibility is to do marching pointers, but instead of incrementing pointers one-by-one to find all intersections, we use binary search to quickly find the next intersection. This mirrors the implementation of our baseline \Cref{alg:sparsevectorinnerprod}. We maintain an index $k_x$ in $\nzlist{\query_j}$ and an index $k_K$ in $\nzlist{\weightmatrixchunk{i}}$. At any given time, we either have $k_x = k_K$, in which case, we emit $\queryscalar_{jk}$ and $\weightmatrixchunk{k,i}$; if $k_x < k_K$, we use a binary search to find the index $k_K'$ in $\nzlist{\weightmatrixchunk{i}}$ where $\nzlist{\query_j}[k_x]$ would be inserted into $\nzlist{\weightmatrixchunk{i}}$ and set $k_K \gets k_K'$; and we handle $k_x > k_K$ similarly. \label{enum:binsearch}
    \item \textbf{Hash-map}: The third possibility is to enable fast random access to the rows of $\weightmatrixchunk{i}$ via a hash-map. The hash-map maps indices $i$ to nonzero rows of $\weightmatrixchunk{i}$. One can iterate over $\queryscalar_{jk}$ for $k \in \nzlist{\query_j}$ and perform a hash-map lookup for each $k$ to retrieve $\weightmatrixchunk{k,i}$ if nonzero. This method is implemented by \textsc{NapkinXC} \cite{jasinska2020probabilistic} for online inference. However, in \textsc{NapkinXC}, it is implemented on a per-column basis, which introduces a massive memory overhead. Matrix chunking significantly reduces this memory overhead. \label{enum:hash}
    \item \textbf{Dense Lookup}: The last possibility is to accelerate the above hash-map access by copying the contents of the hash-map into a dense array of length $\featuredim$. Then, a random access to a row of $\weightmatrixchunk{i}$ is done by an array lookup. This dense array is recycled across the entire program, so afterwards, the dense array must be cleared. This is the method implemented by \textsc{Parabel} \cite{prabhu2018parabel} and \textsc{Bonsai} \cite{khandagale2020bonsai}. \label{enum:dense}
\end{enumerate}

\begin{algorithm}
   \caption{Sparse Vector Chunk Product}
   \label{alg:sparsevectorchunkprod}
\begin{algorithmic}[1]
   \STATE {\bfseries Input:} Sparse row vector $\query \in \Rn{\featuredim}$ and chunk $\chunk \in \Rn{\featuredim \times s}$
   \STATE {\bfseries Output:} The dense product $\query \chunk \in \Rn{s}$
   \STATE Initialize dense result vector: $\vc{z} \gets \vc{0} \in \Rn{s}$
   \STATE \emph{Note}: in the following loop, use one of the iterators as described in items 1 through 4.
   \STATE \emph{Note}: For vector $\star$, $\nzlist{\star}$ denotes the array of indices of nonzeros in $\star$
   \STATE \emph{Note}: For chunk $\ast$, $\nzlist{\ast}$ denotes the array of indices of nonzero rows in $\ast$
   \FOR{scalar $\queryscalar_i$, row $\chunk_i$, where $i \in \nzlist{\query} \cap \nzlist{\chunk}$}
     \FOR{All nonzeros in chunk row: $k \in \nzlist{\chunk_{i}}$}
   \STATE $\vc{z}_k \gets \vc{z}_k + \queryscalar_{i} \chunkscalar_{i, k}$
   \ENDFOR
   \ENDFOR
   \STATE {\bfseries return} $\vc{z}$
\end{algorithmic}
\end{algorithm}

\begin{algorithm}
   \caption{Evaluating Masked Matrix Products}
   \label{alg:maskedmatrixprod}
\begin{algorithmic}[1]
   \STATE {\bfseries Input:} Mask matrix $\maskmatrix \in \{0, 1\}^{\querycount \times \labelcount_\layer}$ in CSR format, query matrix $\querymatrix \in \Rn{\querycount \times \featuredim}$ in CSR format, and weight matrix $\weightmatrix \in \Rn{\featuredim \times \labelcount_\layer}$ in chunked format.
   \STATE {\bfseries Output:} Ranker activation matrix $\actmatrix \in \Rn{\querycount \times \labelcount_\layer}$ in CSR format.
   \STATE Allocate memory for $\actmatrix$ using the sparsity pattern of $\maskmatrix$
   \STATE Initialize $\actmatrix \gets \mat{0}$.
   \STATE Collect nonzero blocks: $A \gets \{(i, j) \mid \maskmatrix^{(i, j)} \neq \mat{0}\}$
   \IF{$\querycount > 1$}
   \STATE Sort nonzero blocks $(i, j) \in A$ by chunk index $j$
   \ENDIF
   \FOR{$(i, j) \in A$}
   \STATE Compute $\actmatrix^{(i, j)} \gets \query_i \weightmatrixchunk{j}$ via \Cref{alg:sparsevectorchunkprod}
   \ENDFOR
      \STATE {\bfseries return} $\actmatrix$.
\end{algorithmic}
\end{algorithm}

There is one final optimization that we have found particularly helpful in reducing inference time --- and that is evaluating the nonzero blocks $\actmatrix^{(j, i)}$ in order of column chunk $i$. Doing this ensures that a single chunk $\weightmatrixchunk{i}$ ideally only has to enter the cache once for all the nonzero blocks $\actmatrix^{(j, i)}$ whose values depend on $\weightmatrixchunk{i}$.

With these optimizations in place, we have found in our performance benchmarks that the hash-map iteration scheme tends to perform best on small to medium sized batches. On large batch sizes, dense lookup performs the best --- this is because we incur the cost of copying the hash-map contents into a dense array only once per chunk if we perform evaluations in chunk order. This cost is then amortized by the increase in the speed of random accesses to the rows of $\weightmatrixchunk{i}$.

To help readers decide which implementation would suit them best, we present a rule of thumb to deciding which iterator to use in \Cref{sec:selectingmethod}. A more detailed examination of time complexity and memory overhead is given in \Cref{tab:complexity}.

\begin{figure*}
\centering
\includegraphics[width=2.1in]{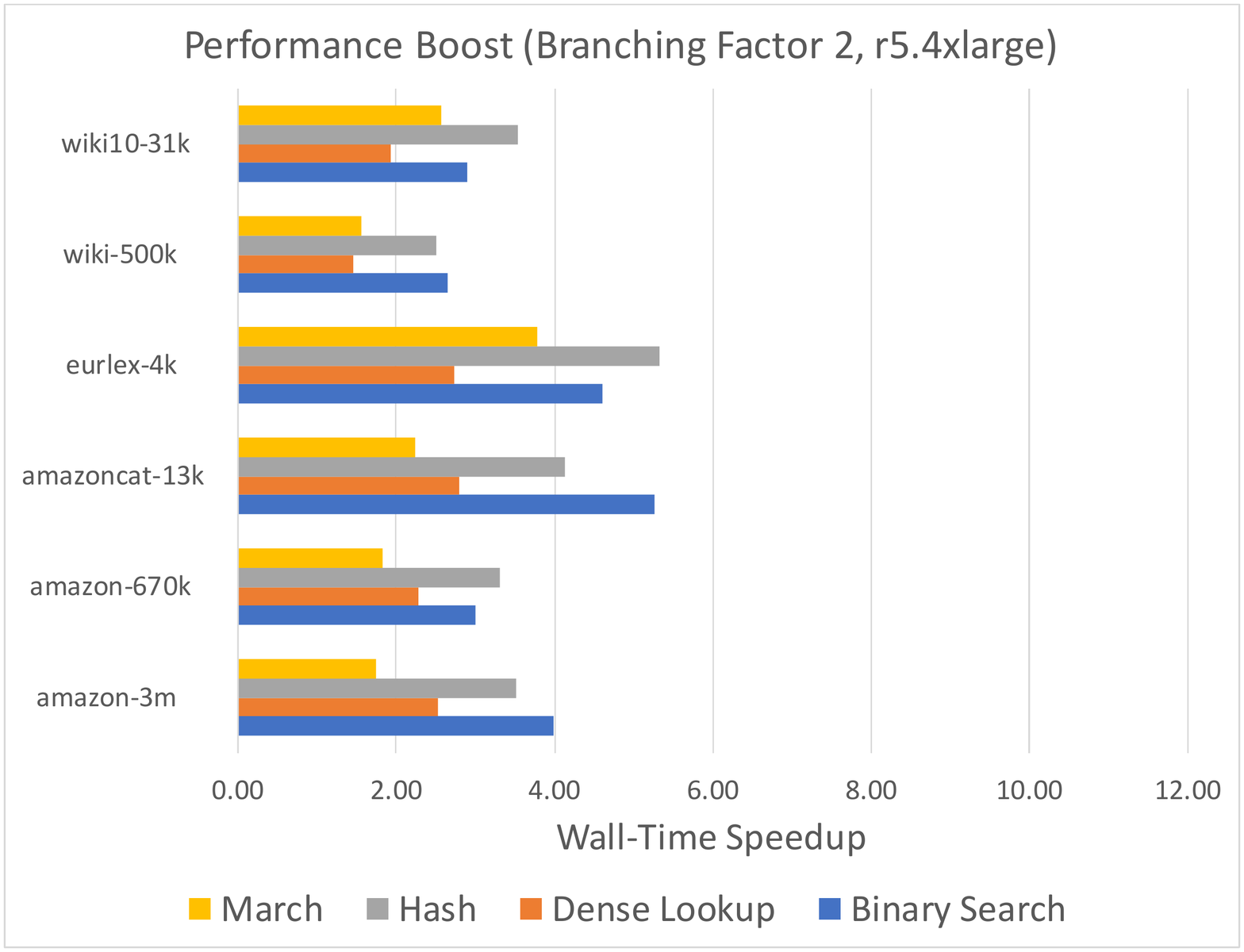}
\includegraphics[width=2.1in]{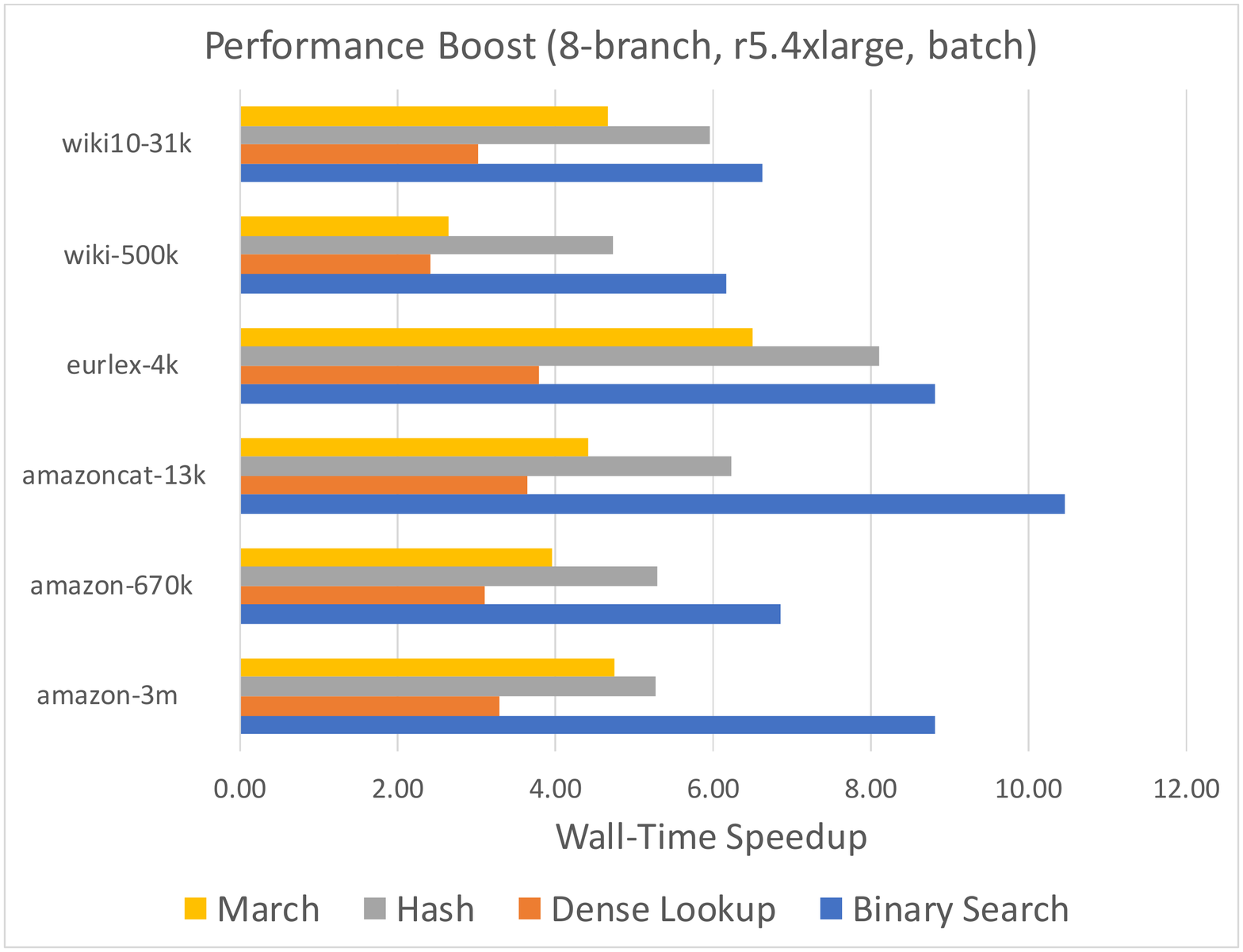}
\includegraphics[width=2.1in]{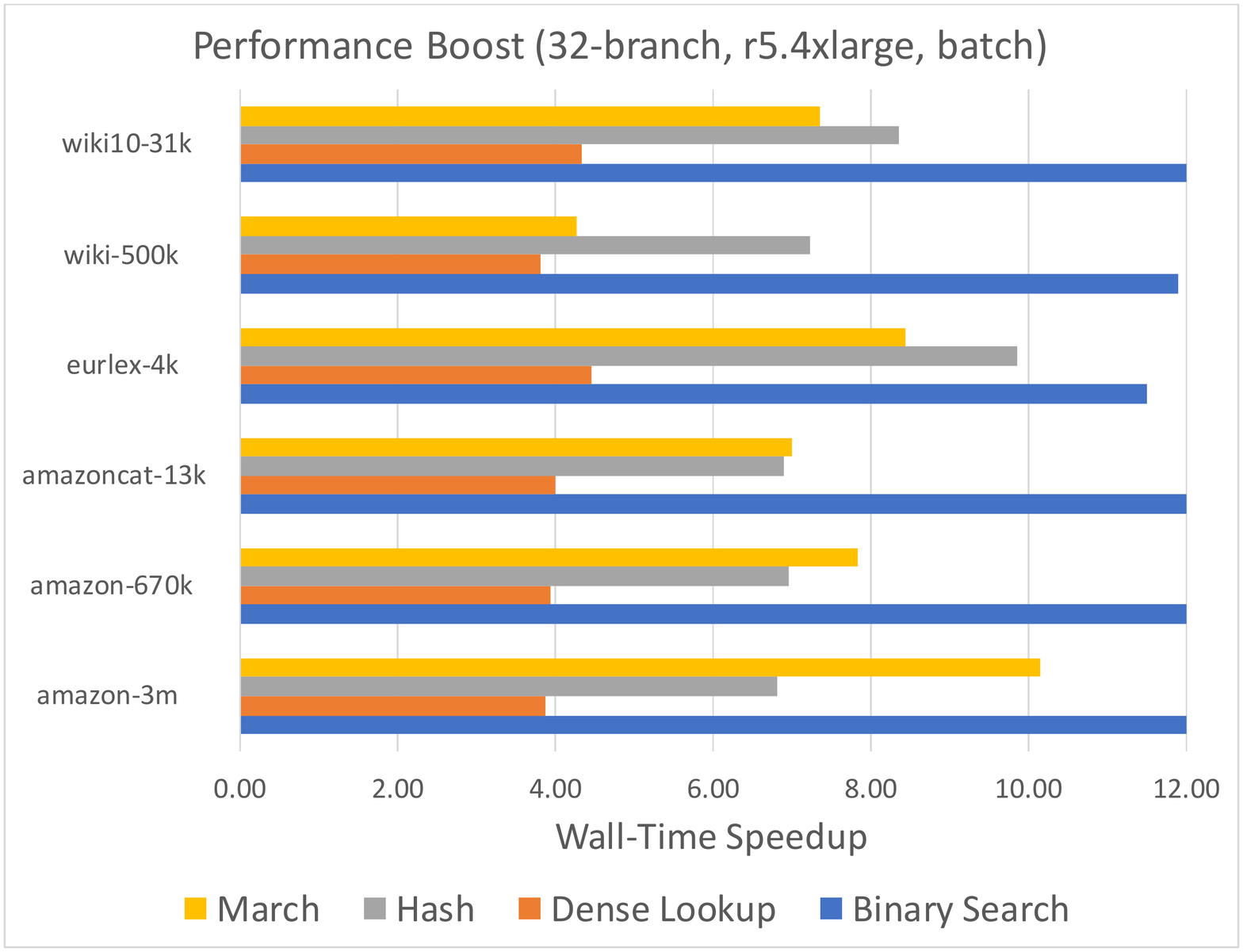}
\caption{A comparison of the speedup provided by an MSCM implementation over a non-MSCM reference implementation for different iteration methods and different tree branching factors in the batch setting. Measured on a \textbf{r5.4xlarge} AWS instance.} \label{fig:inftime2}
\end{figure*}

\begin{figure*}
\centering
\includegraphics[width=2.1in]{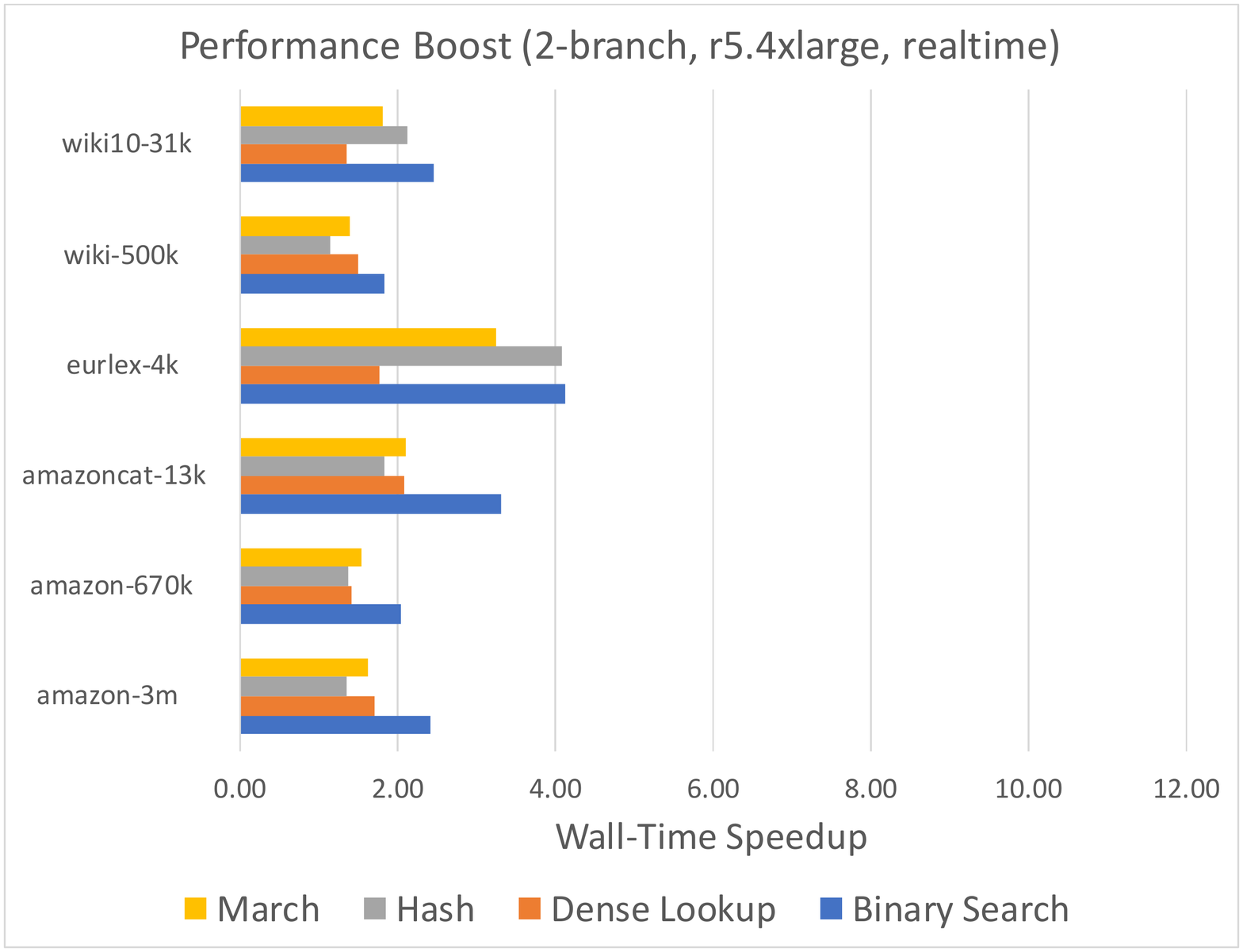}
\includegraphics[width=2.1in]{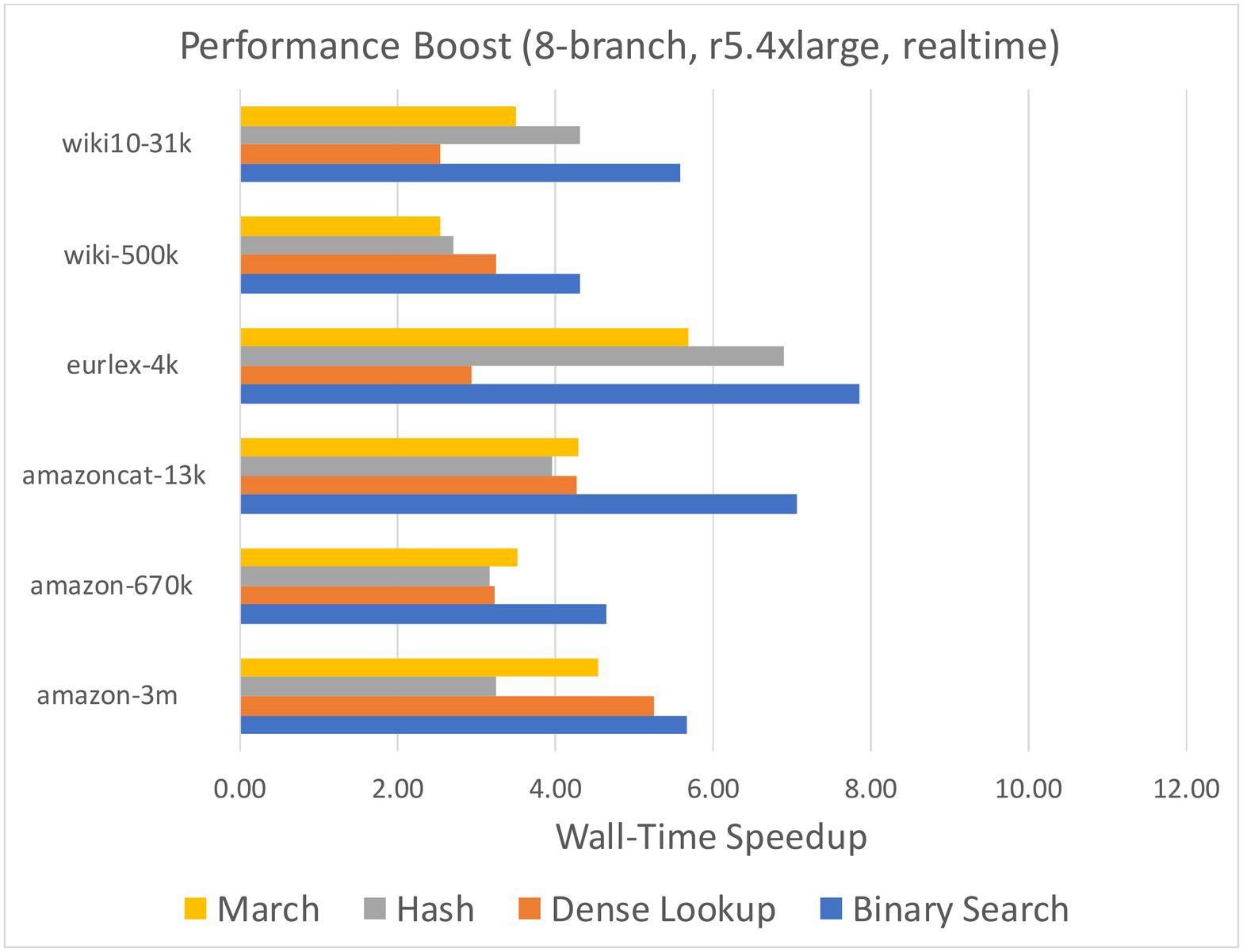}
\includegraphics[width=2.1in]{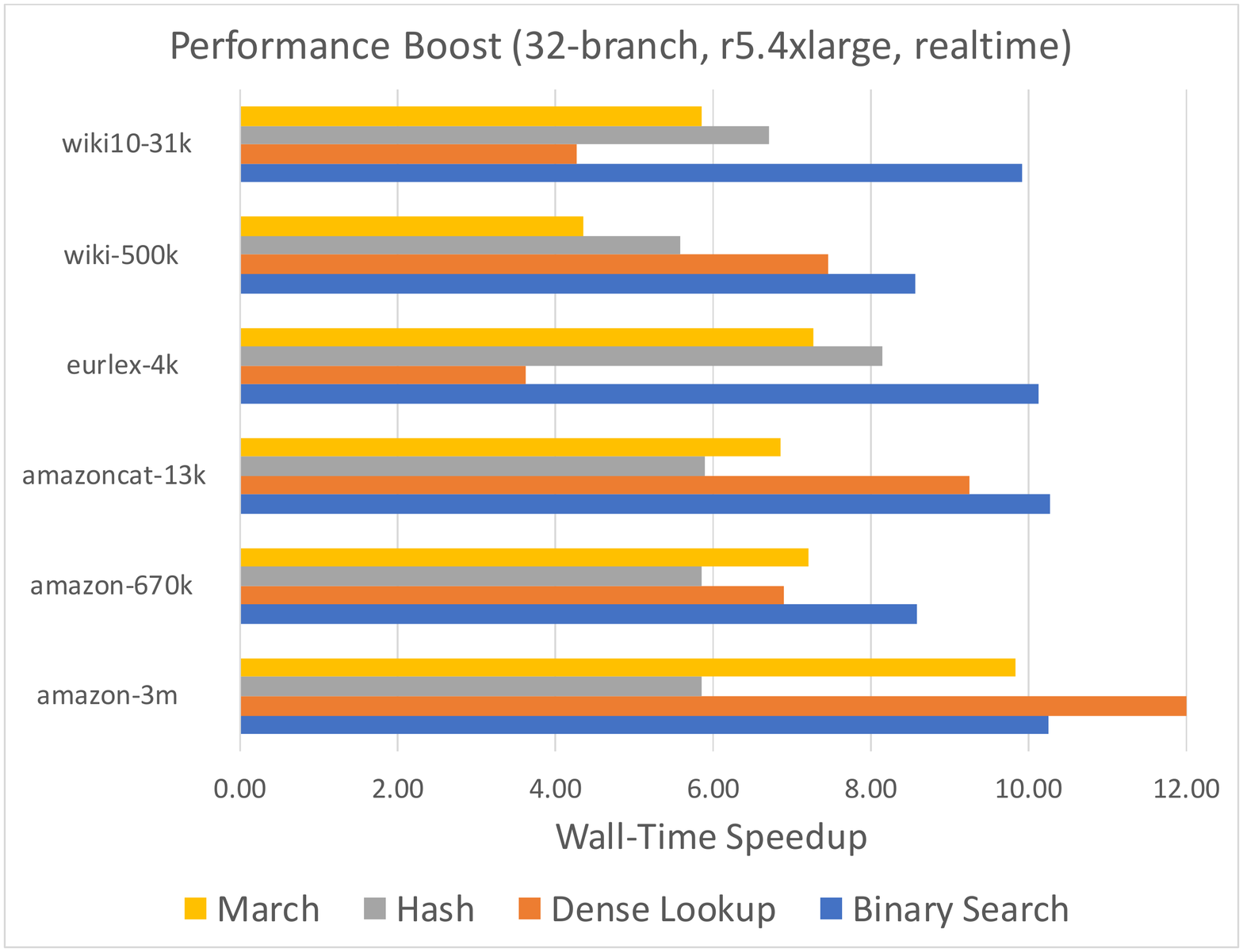}
\caption{A comparison of the speedup provided by an MSCM implementation over a non-MSCM reference implementation for different iteration methods and different tree branching factors in the online setting (i.e., batch size 1). Measured on a \textbf{r5.4xlarge} AWS instance.} \label{fig:inftime3}
\end{figure*}

With all of this taken into account, we present \Cref{alg:maskedmatrixprod}, the final version of our algorithm for performing the masked sparse matrix multiplication in \Cref{eq:maskedprediction}. As one can see in our benchmarks in Figures~\ref{fig:inftime2} and \ref{fig:inftime3}, we have found that the optimizations above produce substantial speedups over the baseline implementation in Alg.~\ref{alg:sparsevectorinnerprod}. We will discuss this in \Cref{sec:eval}. Generally, the larger the branching factor of the model tree, the more significant the performance boost due to the fact that our method aggregates computation over entire chunks. Moreover, we note that the {\bf performance boost of this technique is essentially free} in that it gives exactly the same inference result as the baseline method, but runs significantly faster.

\section{Benchmarks} \label{sec:eval}

To evaluate the extent to which the MSCM technique improves performance, we put all of the above variations of MSCM through a series of benchmarks on a set of models that we trained on six publicly available datasets \cite{Bhatia16}. We ran all benchmarks in this paper on both a reserved \textbf{r5.4xlarge} AWS instance and a reserved \textbf{m5.8xlarge} AWS instance. Results were consistent across both instance types, so in the interest of space, we will present only the \textbf{r5.4xlarge} results here, \textbf{m5.8xlarge} are provided in our benchmark repository.\footnote{https://github.com/UniqueUpToPermutation/pecos/tree/benchmark} The following subsection discusses single-threaded results, while \Cref{sec:parallel} discusses multi-threaded results.

For each dataset, we trained multiple models with a term frequency-inverse document frequency (TFIDF) word embedding and positive instance feature aggregation (PIFA) for label representations (see \cite{anonymous} for more details). In our results we present benchmarks for models constructed with tree branching factors of 2, 8, and 32 to cover a broad swath of performance under different model meta-parameter choices. Furthermore, we ran each of our methods in both the batch setting, where we pass all the queries to the model at the same time as a single matrix, as well as the online setting, where we pass the queries to the model individually one-by-one for evaluation. Both of these settings represent different use cases in industry and we believe it is important for our technique to work well in both settings.
For each choice of model branching factor, setting, dataset, and iteration method (i.e., marching pointers, binary search, hash, dense lookup), we run the inference computation \emph{with MSCM} and \emph{without MSCM} and record the wall-times and the relative speedup that MSCM provides.

For each iteration method, we present the wall-time ratio between an MSCM implementation using that iteration method and a baseline vector-dot-product implementation using that same iteration method. In the vector-dot-product baseline, all entries of the required masked matrix product are computed by taking the inner product between respective rows and columns in the two matrices that form the product.

The results of our benchmarks can be seen in tabular form in \Cref{tab:time_split2}, \Cref{tab:time_split8}, and \Cref{tab:time_split32}. We also provide a visual graph of the inference times relative to our baseline in \Cref{fig:inftime2} and \Cref{fig:inftime3}.

\begin{table*}[p]
  \centering
  { \scriptsize
  \begin{tabular}{lrrrrrr}
    \toprule
      Branching Factor: 2 & amazon-3m & amazon-670k & amazoncat-13k & eurlex-4k & wiki-500k & wiki10-31k \\
    \midrule
    \midrule
    \textbf{Batch} & & & & & & \\ \hdashline
    Binary Search MSCM      & 0.83 \text{ ms} & 0.69 \text{ ms} & 0.45 \text{ ms} & 0.55 \text{ ms} & 3.46 \text{ ms} & 2.69 \text{ ms} \\
    Binary Search           & 2.83 \text{ ms} & 2.02 \text{ ms} & 2.24 \text{ ms} & 2.55 \text{ ms} & 8.75  \text{ ms} & 7.79 \text{ ms} \\ \hdashline
    Dense Lookup MSCM       & 0.43 \text{ ms} & 0.20 \text{ ms} & 0.14 \text{ ms} & 0.18 \text{ ms} & 1.34 \text{ ms} & 0.62 \text{ ms} \\
    Dense Lookup            & 0.82 \text{ ms} & 0.36 \text{ ms} & 0.28 \text{ ms} & 0.49 \text{ ms} & 1.44 \text{ ms} & 1.18 \text{ ms} \\ \hdashline
    Hash MSCM               & 0.49 \text{ ms} & 0.29 \text{ ms} & 0.19 \text{ ms} & 0.25 \text{ ms} & 1.41 \text{ ms} & 0.99 \text{ ms} \\
    Hash                    & 1.26 \text{ ms} & 0.83 \text{ ms} & 0.60 \text{ ms} & 1.34 \text{ ms} & 3.15 \text{ ms} & 3.44 \text{ ms} \\ \hdashline
    Marching Pointers MSCM  & 7.78 \text{ ms} & 2.62 \text{ ms} & 2.88 \text{ ms} & 0.49 \text{ ms} & 16.00 \text{ ms} & 2.13 \text{ ms} \\
    Marching Pointers       & 13.20 \text{ ms} & 4.65 \text{ ms} & 6.31 \text{ ms} & 1.85 \text{ ms} & 24.40 \text{ ms} & 5.40 \text{ ms} \\ 
    \midrule
    \textbf{Online} & & & & & & \\ \hdashline
Binary Search MSCM      & 1.91 \text{ ms} & 1.33 \text{ ms} & 0.95 \text{ ms} & 0.64 \text{ ms} & 7.07 \text{ ms} & 3.27 \text{ ms} \\
Binary Search           & 4.62 \text{ ms} & 2.72 \text{ ms} & 3.16 \text{ ms} & 2.62 \text{ ms} & 12.90 \text{ ms} & 8.06 \text{ ms} \\  \hdashline
Dense Lookup MSCM       & 17.10 \text{ ms} & 4.45 \text{ ms} & 4.87 \text{ ms} & 0.77 \text{ ms} & 74.70 \text{ ms} & 3.04 \text{ ms} \\
Dense Lookup            & 9.10 \text{ ms} & 6.26 \text{ ms} & 10.10 \text{ ms} & 1.37 \text{ ms} & 112.00 \text{ ms} & 4.13 \text{ ms} \\  \hdashline
Hash MSCM               & 1.81 \text{ ms} & 1.32 \text{ ms} & 0.88 \text{ ms} & 0.42 \text{ ms} & 6.20 \text{ ms} & 2.57 \text{ ms} \\ 
Hash                    & 2.44 \text{ ms} & 1.82 \text{ ms} & 1.62 \text{ ms} & 1.73 \text{ ms} & 7.12 \text{ ms} & 5.43 \text{ ms} \\  \hdashline
Marching Pointers MSCM  & 9.10 \text{ ms} & 3.54 \text{ ms} & 3.52 \text{ ms} & 0.58 \text{ ms} & 20.40 \text{ ms} & 3.12 \text{ ms} \\ 
Marching Pointers       & 14.80 \text{ ms} & 5.48 \text{ ms} & 7.37 \text{ ms} & 1.88 \text{ ms} & 28.20 \text{ ms} & 5.67 \text{ ms} \\
    \bottomrule
  \end{tabular}
  }
  \caption{Inference time per query in both the batch and online settings. The performance results were run on an \textbf{r5.4xlarge} AWS instance, on models trained with a branching factor of 2. For the online results, we test only a random subset of the test data of size 10,000.}
  \label{tab:time_split2}
\end{table*}

\begin{table*}[p]
   \centering
   { \scriptsize
   \begin{tabular}{lrrrrrr}
     \toprule
       Branching Factor: 8 & amazon-3m & amazon-670k & amazoncat-13k & eurlex-4k & wiki-500k & wiki10-31k \\
     \midrule
    \midrule
    \textbf{Batch} & & & & & & \\ \hdashline
   Binary Search MSCM      & 0.43 \text{ ms} & 0.34 \text{ ms} & 0.23 \text{ ms} & 0.29 \text{ ms} & 1.67 \text{ ms} & 1.23 \text{ ms} \\
Binary Search           & 3.01 \text{ ms} & 2.20 \text{ ms} & 2.22 \text{ ms} & 2.52 \text{ ms} & 9.73 \text{ ms} & 8.16 \text{ ms} \\ \hdashline
Dense Lookup MSCM       & 0.32 \text{ ms} & 0.16 \text{ ms} & 0.11 \text{ ms} & 0.13 \text{ ms} & 0.83 \text{ ms} & 0.41 \text{ ms} \\ 
Dense Lookup            & 0.84 \text{ ms} & 0.38 \text{ ms} & 0.28 \text{ ms} & 0.48 \text{ ms} & 1.54 \text{ ms} & 1.21 \text{ ms} \\ \hdashline
Hash MSCM               & 0.33 \text{ ms} & 0.19 \text{ ms} & 0.13 \text{ ms} & 0.16 \text{ ms} & 0.87 \text{ ms} & 0.60 \text{ ms} \\
Hash                    & 1.31 \text{ ms} & 0.87 \text{ ms} & 0.60 \text{ ms} & 1.33 \text{ ms} & 3.36 \text{ ms} & 3.57 \text{ ms} \\ \hdashline
Marching Pointers MSCM  & 3.41 \text{ ms} & 1.34 \text{ ms} & 1.48 \text{ ms} & 0.28 \text{ ms} & 10.10 \text{ ms} & 1.24 \text{ ms} \\
Marching Pointers       & 15.50 \text{ ms} & 5.14 \text{ ms} & 6.35 \text{ ms} & 1.84 \text{ ms} & 26.20 \text{ ms} & 5.79 \text{ ms} \\
    \midrule
    \textbf{Online} & & & & & & \\ \hdashline
Binary Search MSCM      & 0.83 \text{ ms} & 0.59 \text{ ms} & 0.43 \text{ ms} & 0.33 \text{ ms} & 3.25 \text{ ms} & 1.49 \text{ ms} \\
Binary Search           & 4.71 \text{ ms} & 2.76 \text{ ms} & 3.03 \text{ ms} & 2.57 \text{ ms} & 14.00 \text{ ms} & 8.31 \text{ ms} \\ \hdashline
Dense Lookup MSCM       & 6.23 \text{ ms} & 2.04 \text{ ms} & 2.18 \text{ ms} & 0.43 \text{ ms} & 33.60 \text{ ms} & 1.68 \text{ ms} \\
Dense Lookup            & 32.70 \text{ ms} & 6.57 \text{ ms} & 9.32 \text{ ms} & 1.27 \text{ ms} & 109.00 \text{ ms} & 4.28 \text{ ms} \\ \hdashline
Hash MSCM               & 0.79 \text{ ms} & 0.60 \text{ ms} & 0.40 \text{ ms} & 0.24 \text{ ms} & 2.88 \text{ ms} & 1.27 \text{ ms} \\
Hash                    & 2.55 \text{ ms} & 1.89 \text{ ms} & 1.58 \text{ ms} & 1.68 \text{ ms} & 7.77 \text{ ms} & 5.48 \text{ ms} \\ \hdashline
Marching Pointers MSCM  & 3.81 \text{ ms} & 1.69 \text{ ms} & 1.71 \text{ ms} & 0.32 \text{ ms} & 11.90 \text{ ms} & 1.70 \text{ ms} \\
Marching Pointers       & 17.30 \text{ ms} & 5.93 \text{ ms} & 7.35 \text{ ms} & 1.84 \text{ ms} & 30.30 \text{ ms} & 5.96 \text{ ms} \\
    \bottomrule
   \end{tabular}
   }
   \caption{Inference time per query in both the batch and online settings. The performance results were run on an \textbf{r5.4xlarge} AWS instance, on models trained with a branching factor of 8. For the online results, we test only a random subset of the test data of size 10,000.}
   \label{tab:time_split8}
 \end{table*}

 \begin{table*}[p]
   \centering
   { \scriptsize
   \begin{tabular}{lrrrrrr}
     \toprule
       Branching Factor: 32 & amazon-3m & amazon-670k & amazoncat-13k & eurlex-4k & wiki-500k & wiki10-31k \\
     \midrule
    \midrule
    \textbf{Batch} & & & & & & \\ \hdashline
   Binary Search MSCM      & 0.38 \text{ ms} & 0.29 \text{ ms} & 0.21 \text{ ms} & 0.21 \text{ ms} & 1.44 \text{ ms} & 0.95 \text{ ms} \\
Binary Search           & 4.28 \text{ ms} & 3.60 \text{ ms} & 2.88 \text{ ms} & 2.45 \text{ ms} & 16.40 \text{ ms} & 11.60 \text{ ms} \\ \hdashline
Dense Lookup MSCM       & 0.32 \text{ ms} & 0.18 \text{ ms} & 0.12 \text{ ms} & 0.11 \text{ ms} & 0.77 \text{ ms} & 0.37 \text{ ms} \\ 
Dense Lookup            & 0.93 \text{ ms} & 0.54 \text{ ms} & 0.36 \text{ ms} & 0.47 \text{ ms} & 2.34 \text{ ms} & 1.69 \text{ ms} \\ \hdashline
Hash MSCM               & 0.33 \text{ ms} & 0.20 \text{ ms} & 0.14 \text{ ms} & 0.13 \text{ ms} & 0.84 \text{ ms} & 0.53 \text{ ms} \\ 
Hash                    & 1.64 \text{ ms} & 1.19 \text{ ms} & 0.76 \text{ ms} & 1.32 \text{ ms} & 5.04 \text{ ms} & 4.41 \text{ ms} \\ \hdashline
Marching Pointers MSCM  & 2.07 \text{ ms} & 1.06 \text{ ms} & 1.25 \text{ ms} & 0.21 \text{ ms} & 8.78 \text{ ms} & 1.01 \text{ ms} \\ 
Marching Pointers       & 20.00 \text{ ms} & 8.03 \text{ ms} & 8.44 \text{ ms} & 1.78 \text{ ms} & 36.80 \text{ ms} & 7.42 \text{ ms} \\ 
    \midrule
    \textbf{Online} & & & & & & \\ \hdashline
    Binary Search MSCM      & 0.69 \text{ ms} & 0.51 \text{ ms} & 0.37 \text{ ms} & 0.25 \text{ ms} & 2.70 \text{ ms} & 1.18 \text{ ms} \\ 
Binary Search           & 7.06 \text{ ms} & 4.40 \text{ ms} & 3.81 \text{ ms} & 2.51 \text{ ms} & 23.10 \text{ ms} & 11.70 \text{ ms} \\ \hdashline
Dense Lookup MSCM       & 3.57 \text{ ms} & 1.55 \text{ ms} & 1.84 \text{ ms} & 0.33 \text{ ms} & 23.50 \text{ ms} & 1.35 \text{ ms} \\ 
Dense Lookup            & 44.60 \text{ ms} & 10.70 \text{ ms} & 17.00 \text{ ms} & 1.20 \text{ ms} & 175.00 \text{ ms} & 5.77 \text{ ms} \\ \hdashline 
Hash MSCM               & 0.61 \text{ ms} & 0.48 \text{ ms} & 0.33 \text{ ms} & 0.20 \text{ ms} & 2.26 \text{ ms} & 1.06 \text{ ms} \\ 
Hash                    & 3.59 \text{ ms} & 2.80 \text{ ms} & 1.93 \text{ ms} & 1.63 \text{ ms} & 12.60 \text{ ms} & 7.10 \text{ ms} \\ \hdashline
Marching Pointers MSCM  & 2.30 \text{ ms} & 1.28 \text{ ms} & 1.41 \text{ ms} & 0.25 \text{ ms} & 9.98 \text{ ms} & 1.30 \text{ ms} \\ 
Marching Pointers       & 22.60 \text{ ms} & 9.23 \text{ ms} & 9.66 \text{ ms} & 1.78 \text{ ms} & 43.50 \text{ ms} & 7.61 \text{ ms} \\
    \bottomrule
   \end{tabular}
   }
   \caption{Inference time per query in both the batch and online settings. The performance results were run on an \textbf{r5.4xlarge} AWS instance, on models trained with a branching factor of 32. For the online results, we test only a random subset of the test data of size 10,000.}
   \label{tab:time_split32}
 \end{table*}

\subsection{Discussion}

It is evident from the results in Tables~\ref{tab:time_split2}, \ref{tab:time_split8}, \ref{tab:time_split32} and Figures~\ref{fig:inftime2} and \ref{fig:inftime3} that the MSCM technique provides a substantial acceleration to any baseline inference technique. Moreover, in the batch setting with a single thread, we observe that chunked matrices with dense lookup always perform faster than every other technique, regardless of dataset or tree topology. Moreover, this speed boost grows more substantial as the branching factor of the model grows. This matches our expectations, because larger branching factors allow the chunked methods to shave off more of the unnecessary traversal costs and cache misses of the unchunked methods.

In online mode, there is no longer always a clear winner in the performance benchmarks, but it seems that hash-map chunked matrices usually provide optimal or near-optimal performance among the algorithms that we have tested here, although there is more individual variation among the results than in the batch setting. We note that the dense lookup methods tend to perform worse in this setting because the cost of loading a hash-map into a dense vector is no longer amortized across a large number of queries. Furthermore, we also note that, as one would expect, larger branching factors give a more substantial performance increase. These performance gains persist in multi-threaded environments, see \cref{sec:parallel} for details about multi-threaded performance.

From these results, we can conclude that it is \emph{always} beneficial to use a chunked MSCM matrix format over a column-by-column (e.g., CSC) format for the weight matrix.

\subsection{Performance Comparison versus NapkinXC}

\begin{figure}
    \centering
    \includegraphics[width=2.5in]{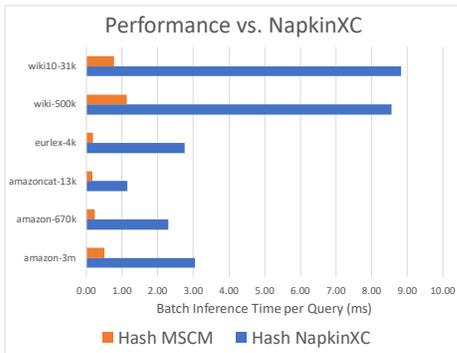}
    \caption{A direct performance comparison between our inference code's performance (with MSCM) to a reference implementation (NapkinXC \cite{jasinska2020probabilistic}). Both implementation use the hash iteration scheme. We see that MSCM enables a nearly $10\times$ performance gain over NapkinXC.}
    \label{fig:napkinxc}
\end{figure}

To help compare the performance of our code-base with MSCM implemented against other state of the art code-bases, we have implemented a direct comparison with NapkinX \cite{jasinska2020probabilistic}. We built a script that converts models in our format to models in NapkinXC's format, allowing us to do an apples-to-apples comparison with an external implementation. We can see in \cref{fig:napkinxc} that we substantially outperform NapkinXC by a margin of $\sim10 \times$ on every dataset. Our conversion script is accessible at \url{https://github.com/UniqueUpToPermutation/PECOStoNapkinXC}.

\section{ Enterprise-Scale Performance} \label{sec:enterprise}

\begin{table*}[t]
  \centering
  { \small 
  \begin{tabular}{l c c c}
    \toprule
      Iteration Method & Average Time (ms/query) & P95 Time (ms/query) & P99 Time (ms/query) \\
      \midrule
      \textit{Beam Size: 10} & \\ 
      \hdashline
      Binary Search MSCM & 0.88 &	1.20 &	1.42 \\
      Hash-map MSCM & 0.96	& 1.38	& 1.62 \\
      Binary Search & 7.28	& 10.56	& 12.78 \\
       \midrule
      \textit{Beam Size: 20} &  \\ 
      \hdashline
      Binary Search MSCM & 1.63	& 2.23	& 2.63	\\
      Hash-map MSCM & 1.80 &	2.60 &	3.06	\\
      Binary Search & 14.32 &	20.68 &	24.81	\\
       \bottomrule \\
  \end{tabular}
  }
  \caption{Performance results on an enterprise-scale semantic search model with 100 million labels in batch mode using a single thread on an \textbf{X1} AWS instance. The model's branching factor is 32. Dense lookup is not compared due to out of memory issue.} \label{tab:enterprise}
\end{table*}

To end our discussion of the performance benefits of MSCM, we deploy our MSCM method on an enterprise-scale semantic search model with $L=100$ million labels (products) and a feature dimension of $d=4$ million. We see that with beam size $10$, both binary search and hash-map MSCM deliver sub-millisecond latency per query on this exceedingly large search problem on 100 million products --- a more than $8\times$ reduction in average and P95/P99 inference times compared to vanilla binary search algorithms. In particular, binary search MSCM with beam size 10 has average inference time of 0.884 ms while binary search without MSCM needs 7.282 ms. In terms of P99 latency, binary search MSCM gains more over binary search without MSCM --- 1.416 ms vs 12.781 ms. More performance results are provided in \Cref{tab:enterprise} in the Appendix. 

\subsection{Multi-Threaded MSCM}\label{sec:parallel}

\begin{figure*}
    \centering
    \includegraphics[width=2.1in]{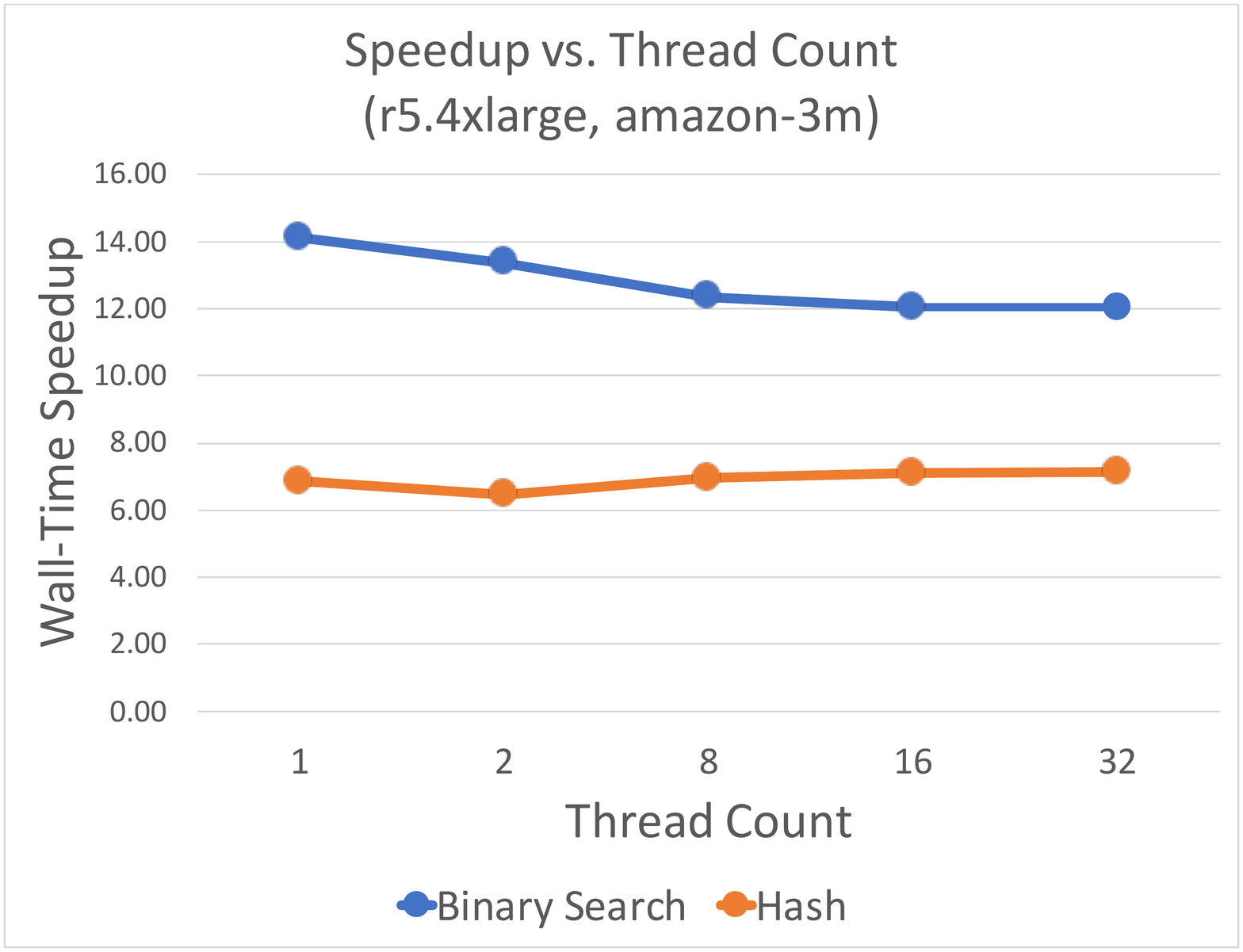}
    \includegraphics[width=2.1in]{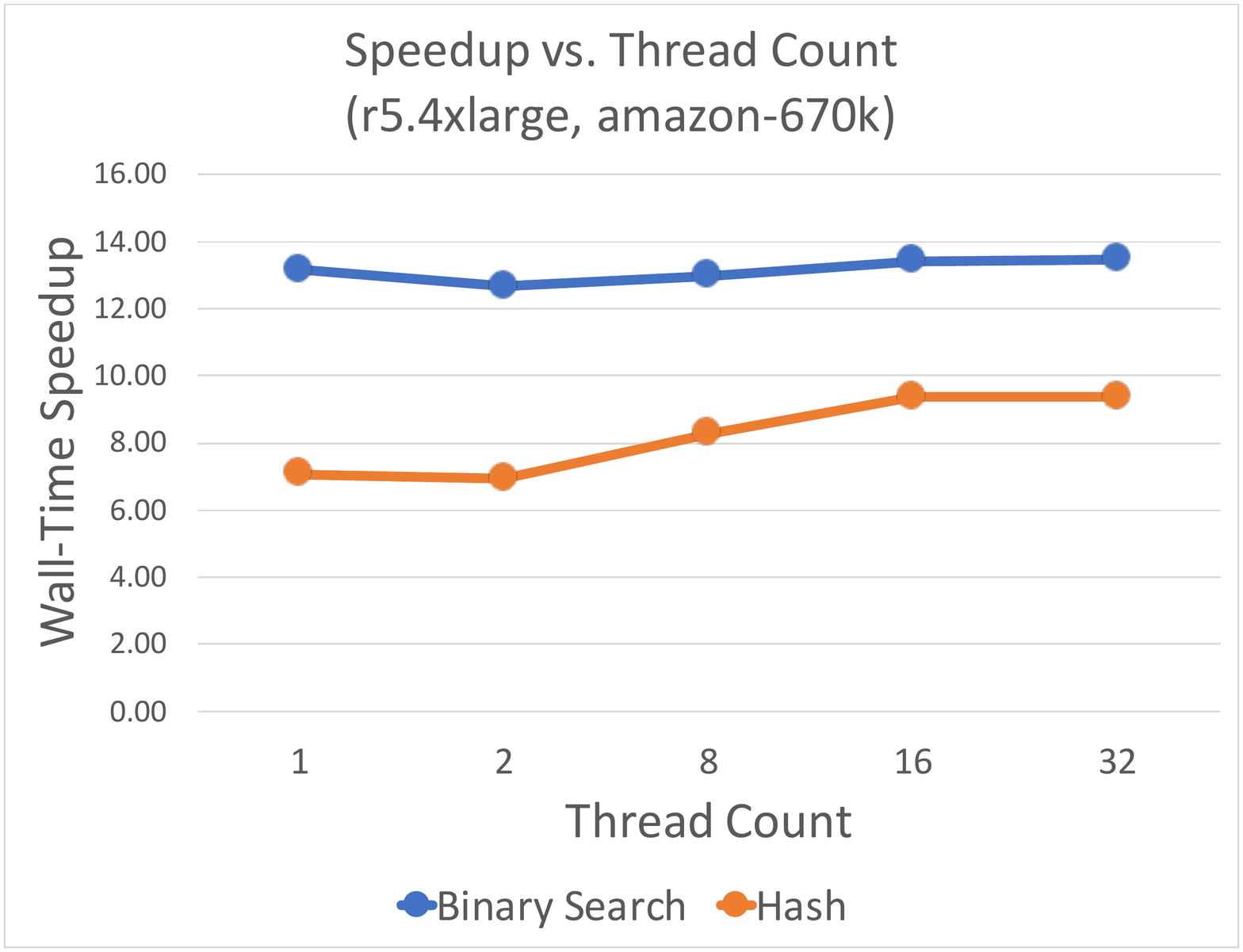}
    \includegraphics[width=2.1in]{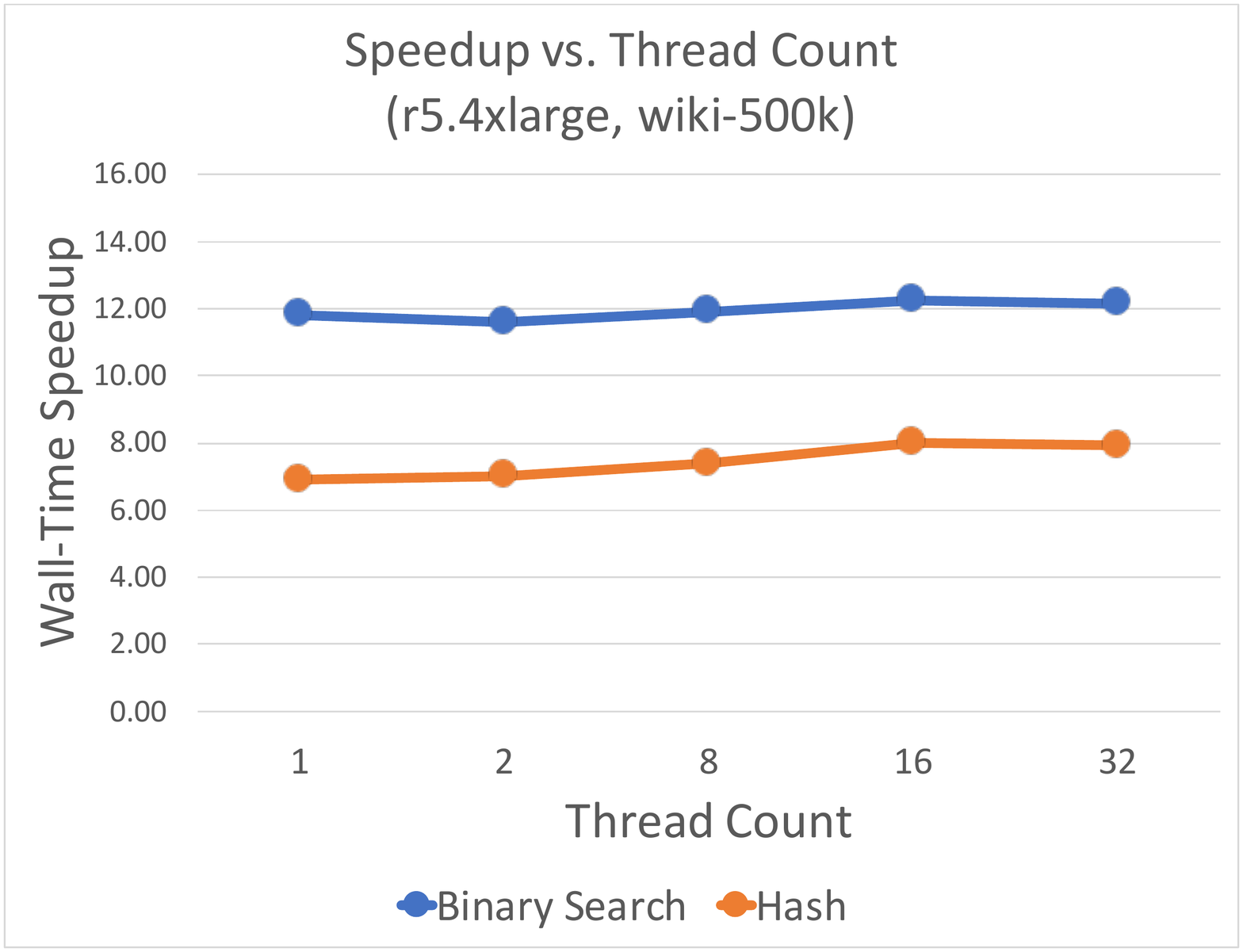}
    \caption{A figure demonstrating the easy parallelization of MSCM. We see that the relative speed-up across different numbers of threads is consistent.}
    \label{fig:multithread}
\end{figure*}

One of the benefits of binary search and hash-map MSCM techniques is that batch processing is embarrassingly parallelizable. Indeed, for binary search and hash-map MSCM, one can simply distribute all of the row-chunk operations in line (7) of \Cref{alg:sparsevectorchunkprod} across many different threads with a library like OpenMP, and no substantial additional work or synchronization is required. Dense lookup (both the baseline and MSCM variants) is harder to parallelize because each thread requires its own dense lookup; moreover, we observe that the performance of dense lookup and its MSCM variant are not competitive when parallelized, due to these subtleties that arise in implementation. Since there are no alternative parallel dense lookup tree-based XMR methods available at time of writing (\textsc{Parabel} and \textsc{Bonsai} do not offer parallelization at this level of granularity), we leave this as possible future work and focus on binary search and hash-map MSCM.

We present the results of parallelizing MSCM in \Cref{fig:multithread}, which clearly shows the performance benefits of MSCM clearly extend into the multi-threaded regime. In each of our three tests on \textsc{wiki-500k}, \textsc{amazon-670k}, and \textsc{amazon-3m}, both binary search and hash-map MSCM are significantly faster than their non-MSCM counterparts.
\section{Conclusion} \label{sec:conclusion}

In this paper, we have presented the MSCM technique for accelerating inference for sparse XMR tree models. MSCM leverages the structure of the inference sparsity pattern to mitigate unnecessary traversal and improve memory locality. MSCM also comes \textbf{free-of-charge} as it exactly preserves inference results, and only requires modest changes to an existing inference pipeline. Moreover, we have examined different variations of the MSCM technique to compile guidelines for extracting the optimal performance. We believe that this technique will allow practitioners to save a substantial amount of compute resources in their inference pipelines, in both online and offline prediction settings. We have already implemented it into our production pipeline and have been very pleased with the low latency our method enables, as demonstrated in \cref{sec:enterprise}.

For future work, another avenue for optimization comes from the observation that a substantial part of our performance boost comes from sorting vector times chunk products by chunk id --- thus better localizing our computation in memory space. It is possible reordering the queries in order to achieve a similar effect. We briefly investigated this, but were unable to obtain a performance boost in our exploration. Further, our exploration of MSCM techniques is not exhaustive, there may be additional ways to iterate through the intersection of query and chunk supports.

As for limitations of the work presented herein, we note that our technique explicitly requires that the XMR tree is composed of linear rankers --- which means our method is not directly applicable to models that use nonlinear based rankers, such as neural networks. However, since most neural network architectures are repeated compositions of linear transformations and activation functions, our technique may be applicable in sparse settings. Also, we note that our technique as presented here is designed to run on the CPU, and one might gain additional performance by investigating GPU-based implementations.
\bibliographystyle{plain}
\bibliography{pecos_inference_main}
\appendix

\section{Appendix}

\begin{table*}[t]
  \centering
  { \scriptsize
  \begin{tabular}{lrrrrrr}
  \toprule
    Dataset & eurlex-4k & amazoncat-13k & wiki10-31k & wiki-500k & amazon-670k & amazon-3m\\ \midrule
    Dimension ($\featuredim$) & 5K & 204K & 102K & 2M & 136K & 337K \\
    Labels ($\labelcount$) & 4K & 13K & 31K & 501K & 670K & 3M \\
    Train Data Size & 16K & 1M & 14K & 2M & 490K & 2M \\
    Test Data Size & 4K & 307K & 7K & 784K & 153K & 743K \\ \bottomrule
    \end{tabular} 
    }
  \caption{Size statistics for all datasets in our experiments}
  \label{tab:data_stats}
  
\end{table*}

\begin{table*}[t]
    \centering
    \begin{tabular}{ccc}
    \toprule
        Iteration Method & Per-Query Time Complexity & Extra Memory Overhead \\
        \midrule
        Marching Pointers & $O(\text{nnz}_x + \text{nnz}_K)$ & None \\
        Binary Search & $O(\min(\text{nnz}_x, \text{nnz}_K) \cdot \log(\max(\text{nnz}_x, \text{nnz}_K))$ & None  \\
        Hash-map & $O(h \cdot \text{nnz}_x)$ & $O(c \cdot \text{nnz}_K)$ \\
        Dense Lookup & $O(\text{nnz}_x + \text{nnz}_K / \querycount)$ & $O(\featuredim)$
        \\ \bottomrule
    \end{tabular}
    \caption{A chart of the per-query time complexity and memory overhead of all the different iteration methods provided in this section. Here $h$ denotes the time it takes to perform a hash lookup, $c$ denotes the number of chunks, $\text{nnz}_x = |\nzlist{\vc{x}}|$ and $\text{nnz}_K = |\nzlist{\mat{K}}|$.}
    \label{tab:complexity}
\end{table*}

\begin{algorithm}[H]
   \caption{Sparse Vector Inner Product}
   \label{alg:sparsevectorinnerprod}
\begin{algorithmic}[1]
   \STATE {\bfseries Input:} Sparse vectors $\query \in \Rn{\featuredim}$ and $\vc{y} \in \Rn{\featuredim}$
   \STATE {\bfseries Output:} The value of $\query \cdot \vc{y}$
   \STATE Initialize result variable: $z \gets 0$.
   \STATE Initialize nonzero entry indices: $i_x \gets 0$ and $i_y \gets 0$
   \STATE \emph{Note}: For vector $\star$, $\nzlist{\star}$ denotes the array of indices of nonzeros in $\star$
   \STATE \emph{Note}: The $\text{LowerBound}(L, i)$ function finds the index of the first element in list $L$ that is not less than $i$.
   \REPEAT
   \STATE Get index of $i_x$-th nonzero in $\query$: $j_x \gets \nzlist{\query}[i_x]$
   \STATE Get index of $i_y$-th nonzero in $\vc{y}$: $j_y \gets \nzlist{\vc{y}}[i_y]$
   \IF{Collision check: $j_x = j_y$}
   \STATE $z \gets z + \queryscalar_{j_x} \vc{y}_{j_y}$
   \STATE Increment $i_x$ and $i_y$.
   \ELSIF{$j_x < j_y$}
   \STATE Advance $i_x$: $i_x \gets \text{LowerBound}(\nzlist{\query}, j_y)$
   \ELSIF{$j_y < j_x$}
   \STATE Advance $i_y$: $i_y \gets \text{LowerBound}(\nzlist{\vc{y}}, j_x)$
   \ENDIF
   \UNTIL{$i_x = \text{Length}(\nzlist{\query})$ and $i_y = \text{Length}(\nzlist{\vc{y}})$}
   \STATE {\bfseries return} $z$
\end{algorithmic}
\end{algorithm}
 
\subsection{Selecting an Iteration Method} \label{sec:selectingmethod}

Using the results of the performance benchmark, as well as the contents of \Cref{tab:complexity}, we briefly provide a guide to choosing a version of the MSCM technique that performs well in a given setting. We always recommend using MSCM, and we suggest that the user consider their choice of iteration scheme in the following order:

\begin{enumerate}
  \item \textbf{Dense Lookup MSCM}: Use this when batch sizes are sufficiently large and storing a dense vector of feature dimension $\featuredim$ is not an issue. Conversely, when the batch size is small (i.e., online settings), we don't recommend using dense lookup.
  \item \textbf{Hash-map MSCM}: We recommend using this technique when the queries are significantly sparser than the MSCM chunks. However, this technique typically requires some memory overhead (in our implementation around 40\% additional memory), because a hash-map of nonzero rows must be stored for every chunk.
  \item \textbf{Binary Search MSCM}: We recommend using this if the extra memory requirements of hash-map MSCM are too demanding, as this seems to be a good alternative to hash-maps with only a slight loss in performance, according to our benchmarks. Also, we recommend using this if the weight matrices are significantly sparser than the queries.
  \item \textbf{Marching Pointers MSCM}: We have not found any situations in our data where this outperforms the above two. However, there may be hypothetical cases where marching pointers can perform better than hash-maps or binary search. For example, if queries and chunks are equally sparse, then marching pointers could have the best complexity.
\end{enumerate}


\end{document}